%% file: main.tex
\documentclass[10pt,twocolumn,letterpaper]{article}

\usepackage[pagenumbers]{cvpr} 

\usepackage{graphicx}
\usepackage{amsmath}
\usepackage{amssymb}
\usepackage{booktabs}
\usepackage{mathtools}

\usepackage{times}
\usepackage{epsfig}
\usepackage{amsmath}
\usepackage{bbm}
\DeclareMathAlphabet\mathbfcal{OMS}{cmsy}{b}{n}

\usepackage{float}

\usepackage{lipsum}
\usepackage{stfloats}
\usepackage{multicol}
\usepackage{multirow}
\usepackage{bm}
\usepackage{etoolbox}
\usepackage{icomma}
\usepackage{array}
\usepackage{tabulary}
\usepackage[table]{xcolor}
\usepackage{paralist}
\usepackage{booktabs}
\usepackage{adjustbox}
\usepackage{caption}
\usepackage{subcaption}
\usepackage{arydshln}

\definecolor{gray}{rgb}{0.3,0.3,0.3}
\definecolor{blue}{rgb}{0,0.5,1}
\definecolor{mask_red}{rgb}{1,0,0.8}
\definecolor{green}{rgb}{0.2,1,0.2}
\definecolor{rblue}{rgb}{0,0,1}
\definecolor{lightblue}{HTML}{6495ed}
\definecolor{lightred}{HTML}{F19C99}

\newcommand{\green}[1]{\textcolor[RGB]{96,177,87}{#1}}

\newcommand{\fn}[1]{\footnotesize{#1}}
\newcommand{\gbf}[1]{\green{\bf{\fn{(#1)}}}}

\newcommand{\obf}[1]{\textcolor{orange}{\bf{\fn{(#1)}}}}
\definecolor{graytablerow}{gray}{0.6}

\usepackage[pagebackref,breaklinks,colorlinks]{hyperref}
\hypersetup{colorlinks, citecolor=blue}

\usepackage[capitalize]{cleveref}
\crefname{section}{Sec.}{Secs.}
\Crefname{section}{Section}{Sections}
\Crefname{table}{Table}{Tables}
\crefname{table}{Tab.}{Tabs.}

\begin{document}

\title{Delivering Arbitrary-Modal Semantic Segmentation}

\author{Jiaming Zhang$^{1,}$\thanks{Equal contribution.},
~~Ruiping Liu$^{1,*}$,
~~Hao Shi$^3$,
~~Kailun Yang$^{2,}$\thanks{Corresponding author (e-mail: {\tt kailun.yang@hnu.edu.cn}).},
~~Simon Reiß$^{1}$,\\
~~Kunyu Peng$^1$,
~~Haodong Fu$^4$,
~~Kaiwei Wang$^3$,
~~Rainer Stiefelhagen$^1$\\
\normalsize
$^1$Karlsruhe Institute of Technology,
\normalsize
~$^2$Hunan University,
\normalsize
~$^3$Zhejiang University,
\normalsize
~$^4$Beihang University
}
\maketitle

\begin{abstract}
\input{Tex_content/abstract}
\end{abstract}

\section{Introduction}
\label{sec:intro}

With the explosion of modular sensors, multimodal fusion for semantic segmentation has progressed rapidly recently~\cite{cao2021shapeconv,chen2020sagate,liu2022cmx} and in turn has stirred growing interest to assemble more and more sensors to reach higher and higher segmentation accuracy aside from more robust scene understanding.
However, most works~\cite{hu2019acnet,wu2020depth_adapted,zhuang2021pmf} and multimodal benchmarks~\cite{ha2017mfnet,silberman2012nyuv2,zhang2021issafe} focus on specific sensor pairs, which lack behind the current trend of fusing more and more modalities~\cite{wang2022tokenfusion,broedermann2022hrfuser}, \ie, progressing towards Arbitrary-Modal Semantic Segmentation (AMSS).

\begin{figure}[!t]
	\centering
    \begin{minipage}[t]{.33\columnwidth}
        \includegraphics[width=0.9\columnwidth]{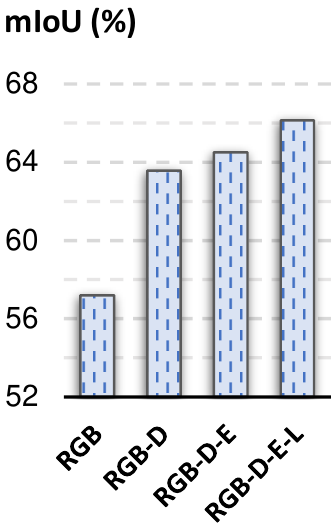}
        \subcaption{RGB-D-E-L fusion.}\label{fig1:rgbdel}
    \end{minipage}%
    \begin{minipage}[t]{.33\columnwidth}
        \includegraphics[width=0.9\columnwidth]{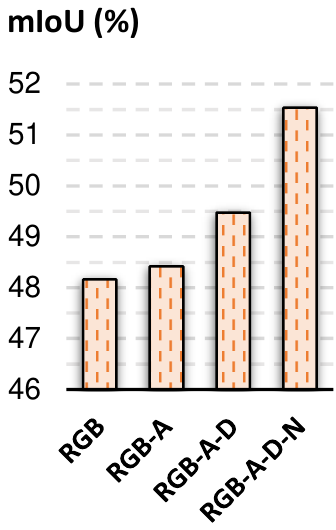}
        \subcaption{RGB-A-D-N fusion.}\label{fig1:rgbadn}
    \end{minipage}%
    \begin{minipage}[t]{.33\columnwidth}
        \includegraphics[width=0.9\columnwidth]{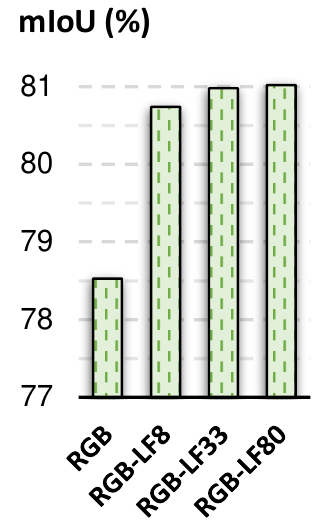}
        \subcaption{RGB-Light Field.}\label{fig1:rgblf}
    \end{minipage}%
    \vskip -1ex
	\caption{Arbitrary-modal segmentation results of CMNeXt using: \\(a).~\{\emph{RGB}, \emph{\textbf{D}epth}, \emph{\textbf{E}vent}, \emph{\textbf{L}iDAR}\} on our \textsc{DeLiVER} dataset; (b).~\{\emph{RGB}, \emph{Angle of Linear Polarization (\textbf{A}oLP)}, \emph{Degree of Linear Polarization (\textbf{D}oLP)}, \emph{Near-Infrared (\textbf{N}IR)}\} on MCubeS~\cite{liang2022mcubesnet}; (c).~\{\emph{RGB}, \emph{8/33/80 sub-aperture Light Fields (LF8/LF33/LF80)} on UrbanLF-Syn~\cite{sheng2022urbanlf}, respectively.}
    \vskip -2ex
    \label{fig1:arbitrary_fusion}
\end{figure}
\begin{figure}[!t]
	\centering
    \begin{minipage}[t]{.42\columnwidth}
        \includegraphics[width=1.0\columnwidth]{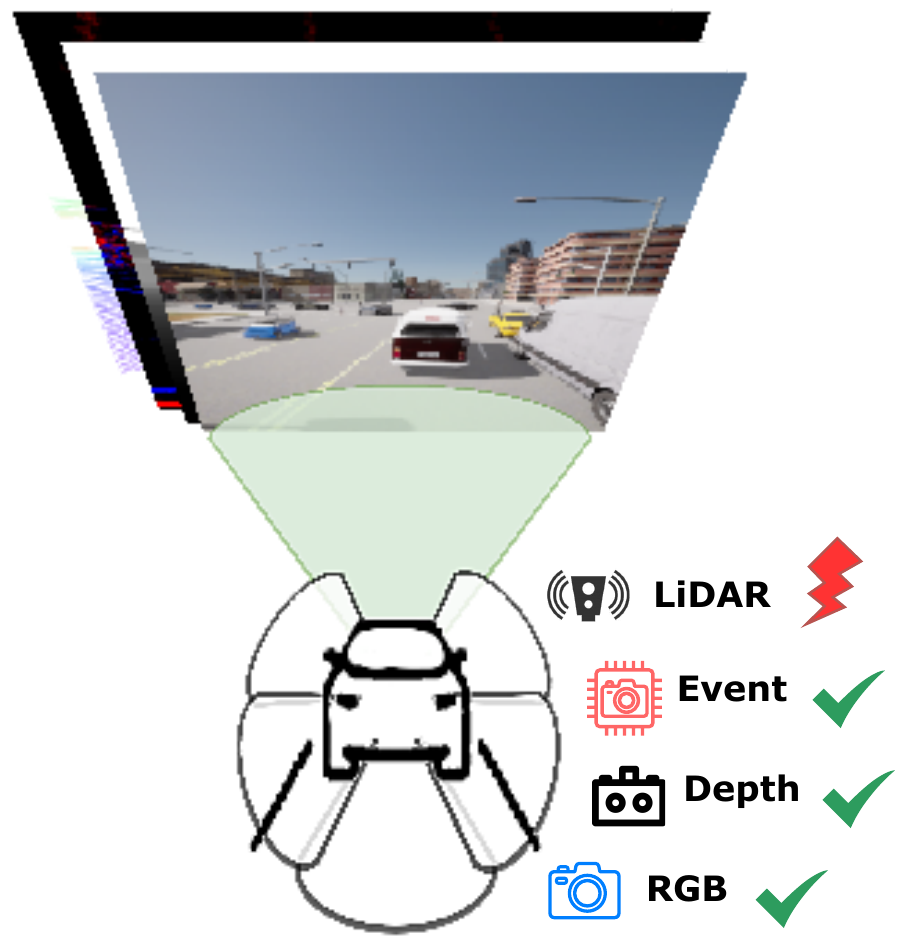}
    \end{minipage}%
    \begin{minipage}[t]{.58\columnwidth}
        \includegraphics[width=1.0\columnwidth]{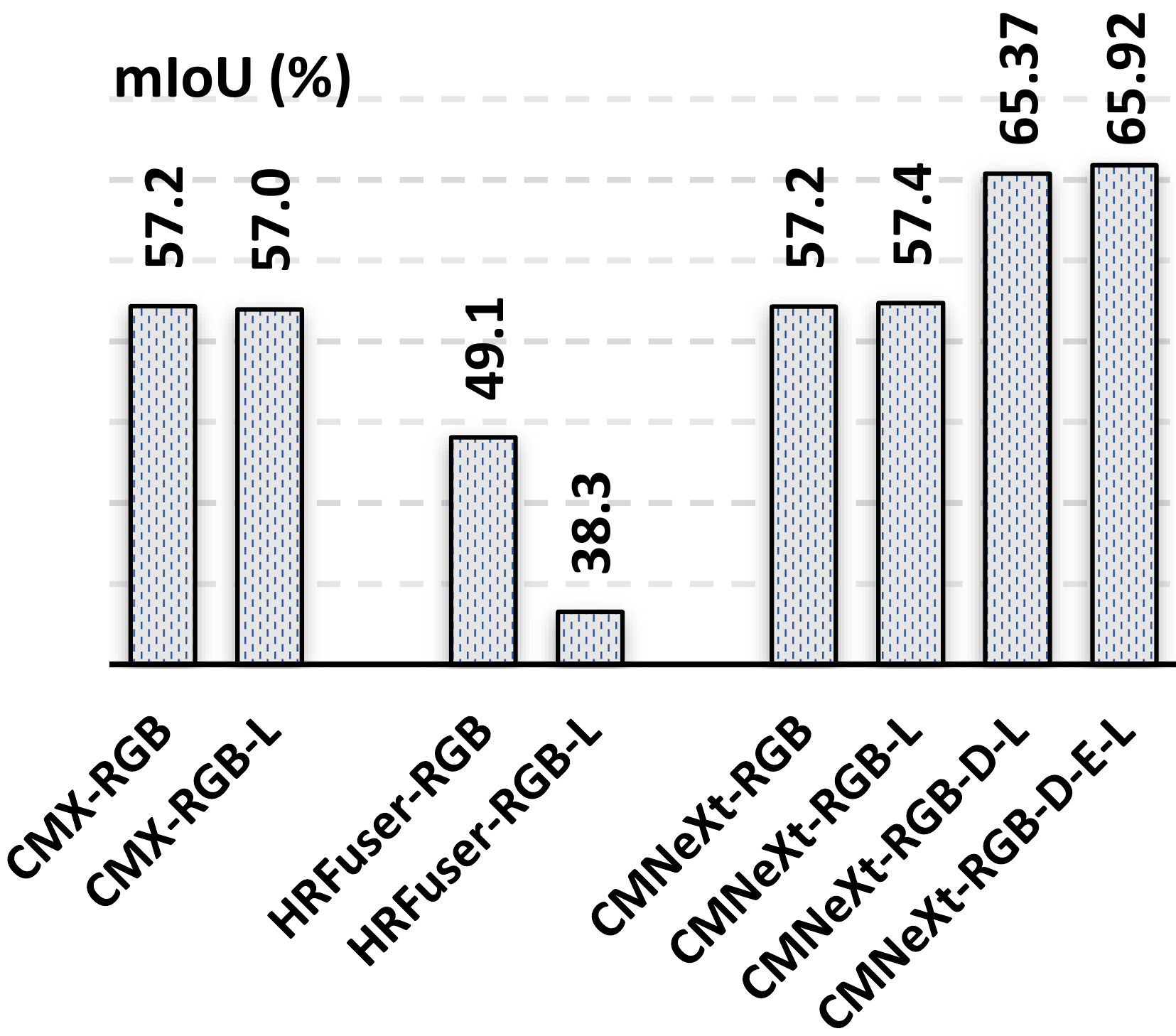}
    \end{minipage}%
    \vskip -2ex
	\caption{Comparing CMX~\cite{liu2022cmx}, HRFuser~\cite{broedermann2022hrfuser}, and our CMNeXt in sensor failure (\ie, LiDAR Jitter) on the \textsc{DeLiVER} dataset.}
    \vskip -2ex
    \label{fig2:sensor_fusion}
\end{figure}

When looking into AMSS, two observations become apparent. Firstly, \emph{an increasing amount of modalities should provide more diverse complementary information,  monotonically increasing segmentation accuracy.}
This is directly supported by our results when incrementally adding and fusing modalities as illustrated in Fig.~\ref{fig1:rgbdel} (RGB-Depth-Event-LiDAR), Fig.~\ref{fig1:rgbadn} (RGB-AoLP-DoLP-NIR), and Fig.~\ref{fig1:rgblf} when adding up to $80$ sub-aperture light-field modalities (RGB-LF$8$/-LF$33$/-LF$80$).
Unfortunately, this great potential cannot be uncovered by previous cross-modal fusion methods~\cite{chen2021sgnet,xiang2021polarization,zhou2021gmnet}, which follow designs for pre-defined modality combinations.
The second observation is that \emph{the cooperation of multiple sensors is expected to effectively combat individual sensor failures.}
Most of the existing works~\cite{wang2020learning_asymmetric_fusion,wu2022transdfusion,valada2020self_multimodal} are built on the assumption that each modality is always accurate.
Under partial sensor faults, which are common in real-life robotic systems,~\eg~LiDAR Jitter, fusing misaligned sensing data might even degrade the segmentation performance, as depicted with CMX~\cite{liu2022cmx} and HRFuser~\cite{broedermann2022hrfuser} in~Fig.~\ref{fig2:sensor_fusion}.
These two critical observations remain to a large extent neglected.

\begin{figure}
    \centering
    \includegraphics[width=1.0\columnwidth]{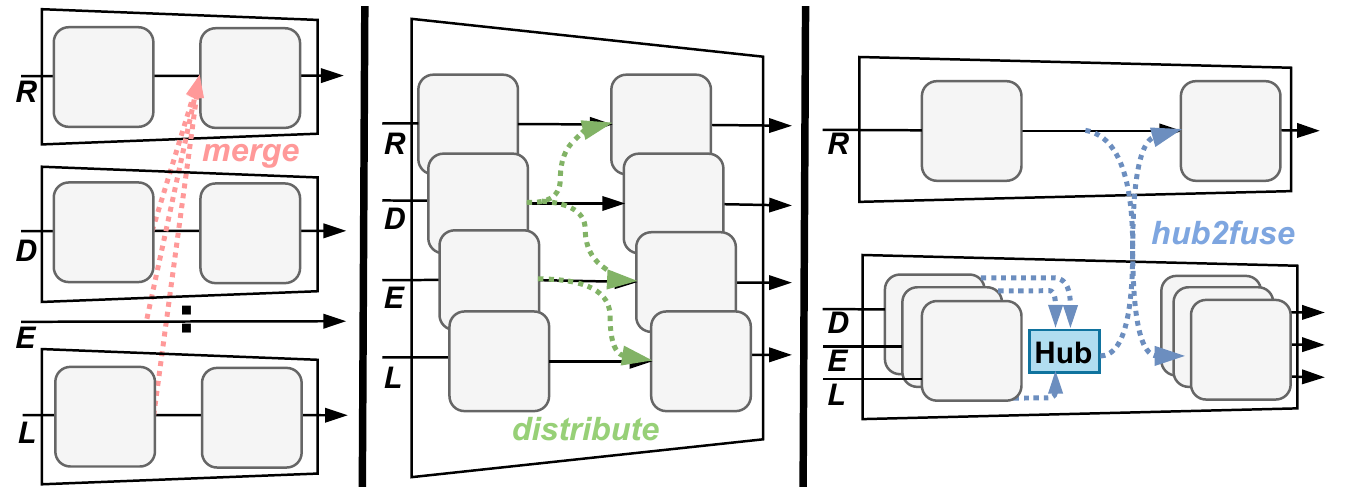}
    \begin{minipage}[t]{.26\columnwidth}
        \vskip -3ex
        \subcaption{Separate}\label{fig3:hrfuser}
    \end{minipage}%
    \begin{minipage}[t]{.36\columnwidth}
        \vskip -3ex
        \subcaption{Joint}\label{fig3:tokenfusion}
    \end{minipage}%
    \begin{minipage}[t]{.36\columnwidth}
        \vskip -3ex
        \subcaption{Asymmetric}\label{fig3:querythenfusion}
    \end{minipage}%
        \vskip -1ex
    \caption{Comparison of multimodal fusion paradigms, such as (a) merging with separate branches~\cite{broedermann2022hrfuser}, (b) distributing with a joint branch~\cite{wang2022tokenfusion}, and (c) our hub2fuse with asymmetric branches.}
    \vskip -2ex
    \label{fig:paradigms}
\end{figure}

To address these challenges, we create a benchmark based on the CARLA simulator~\cite{dosovitskiy2017carla}, with \textbf{De}pth, \textbf{Li}DAR, \textbf{V}iews, \textbf{E}vents, and \textbf{R}GB images: The \textsc{\textbf{DeLiVER}} Multimodal dataset. It features severe weather conditions and five sensor failure modes to exploit complementary modalities and resolve partial sensor outages.
To profit from all this, we present the arbitrary cross-modal \textbf{CMNeXt} segmentation model.
Without increasing the computation overhead substantially when adding more modalities CMNeXt incorporates a novel \emph{Hub2Fuse} paradigm (Fig.~\ref{fig3:querythenfusion}).
Unlike relying on separate branches~(Fig.~\ref{fig3:hrfuser}) which tend to be computationally costly or using a single joint branch~(Fig.~\ref{fig3:tokenfusion}) which often discards valuable information, CMNeXt is an asymmetric architecture with two branches, one for RGB and another for diverse supplementary modalities.

The key challenge lies in designing the two branches to pick up multimodal cues.
Specifically, at the \emph{hub} step of \emph{Hub2Fuse}, to gather useful complementary information from auxiliary modalities, we design a Self-Query Hub~(SQ-Hub), which dynamically selects informative features from all modality-sources before fusion with the RGB branch.
Another great benefit of SQ-Hub is the ease of extending it to an arbitrary number of modalities, at negligible parameters increase~(${\sim}0.01M$ per modality). At the \emph{fusion} step, fusing sparse modalities such as LiDAR or Event data can be difficult to handle for joint branch architectures without explicit fusion such as TokenFusion~\cite{wang2022tokenfusion}. 
To circumvent this issue and make best use of both dense and sparse modalities, we leverage cross-fusion modules~\cite{liu2022cmx} and couple them with our proposed Parallel Pooling Mixer~(PPX) which efficiently and flexibly harvests the most discriminative cues from any auxiliary modality. These design choices come together in our CMNeXt architecture, which paves the way for AMSS (Fig.~\ref{fig1:arbitrary_fusion}). By carefully putting together alternative modalities, CMNeXt can overcome individual sensor failures and enhances segmentation robustness (Fig.~\ref{fig2:sensor_fusion}).

With comprehensive experiments on \textsc{DeLiVER} and five additional public datasets, we gather insight into the strength of the CMNeXt model. On \textsc{DeLiVER}, CMNeXt obtains $66.30\%$ in mIoU with a ${+}9.10\%$ gain compared to the RGB-only baseline~\cite{xie2021segformer}. On UrbanLF-Real~\cite{sheng2022urbanlf} and MCubeS~\cite{liang2022mcubesnet} datasets, CMNeXt surpasses the previous best methods by ${+}3.90\%$ and ${+}8.68\%$, respectively. Compared to previous state-of-the-art methods, our model achieves comparable perfomance on bi-modal NYU Depth V2~\cite{silberman2012nyuv2} as well as MFNet~\cite{ha2017mfnet} and outperforms all previous modality-specific methods on KITTI-360~\cite{liao2022kitti360}.

On a glance, we deliver the following contributions:
\begin{compactitem}
    \item We create the new benchmark \textsc{DeLiVER} for Arbitrary-Modal Semantic Segmentation (AMSS) with four modalities, four adverse weather conditions, and five sensor failure modes.
    \item We revisit and compare different multimodal fusion paradigms and present the \emph{Hub2Fuse} paradigm with an asymmetric architecture to attain AMSS. 
    \item The universal arbitrary cross-modal fusion model CMNeXt is proposed, with a Self-Query Hub~(SQ-Hub) for selecting informative features and a Parallel Pooling Mixer~(PPX) for harvesting discriminative cues.
    \item We investigate AMSS by fusing up to a total of $80$ modalities and notice that CMNeXt achieves state-of-the-art performances on six datasets.
\end{compactitem}

\section{Related Work}
\label{sec:related_work}
\input{Tex_content/related_work}

\begin{figure*}[!t]
	\centering
    \includegraphics[width=1\textwidth]{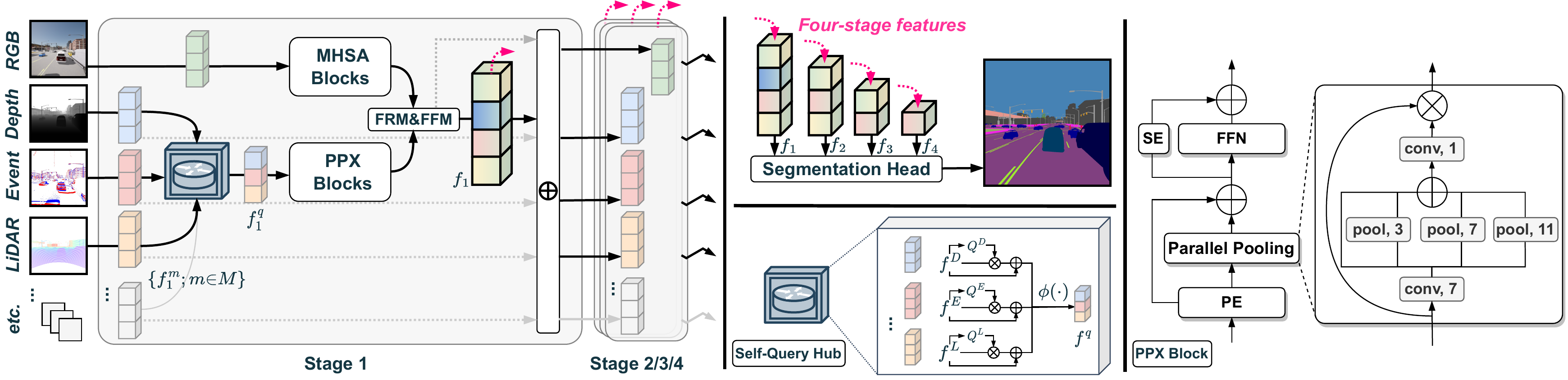}
	\caption{\textbf{CMNeXt architecture} in Hub2Fuse paradigm and asymmetric branches, having \eg Multi-Head Self-Attention (MHSA)~\cite{xie2021segformer} blocks in the RGB branch and our Parallel Pooling Mixer (PPX) blocks in the accompanying branch. At the \emph{hub} step, the Self-Query Hub selects informative features from the supplementary modalities. At the \emph{fusion} step, the feature rectification module (FRM) and feature fusion module (FFM)~\cite{liu2022cmx} are used for feature fusion. Between stages, features of each modality are restored via adding the fused feature. The four-stage \textcolor{magenta}{fused features} are forwarded to the segmentation head for the final prediction. 
	}
    \label{fig:model}
\end{figure*}

\section{CMNeXt: Proposed Framework}
\label{sec:framework}
To achieve arbitrary-modal segmentation, the proposed CMNeXt framework is constructed by using a dual-branch structure in a \emph{Hub2Fuse} paradigm.
We will elaborate the overall CMNeXt architecture in Sec.~\ref{sec:cmnext}, the Self-Query Hub in Sec.~\ref{sec:sq-hub}, and the Parallel Pooling Mixer in Sec.~\ref{sec:ppx}.

\subsection{CMNeXt Architecture}\label{sec:cmnext}
In Fig.~\ref{fig:model}, our CMNeXt has an encoder-decoder architecture.
The encoder is a dual-branch and four-stage encoder. Built on the assumption that the RGB representation is essential for semantic segmentation, the two branches correspond to the primary branch for RGB and the secondary branch for other modalities, respectively.
The four-stage structure follows most of previous CNN/Transformer models~\cite{zhao2017pspnet,fu2019danet,xie2021segformer,wang2021pvt} to extract pyramidal features. 
Note that, Fig.~\ref{fig:model} details only the first of the four stages for brevity.
For the consistency of modal representations, we preprocess LiDAR and Event data as image-like representations following~\cite{zhuang2021pmf,zhang2021issafe}.  
The RGB image $\boldsymbol{I}_{RGB}{\in}H{\times}W{\times}3$ is gradually processed by Multi-Head Self-Attention (MHSA) blocks~\cite{xie2021segformer}, whereas the images of the other $M$ modalities $\boldsymbol{I}_M{\in}H{\times}W{\times}3{\times}M$ by Parallel Pooling Mixer~(PPX) blocks.
After four stages, there are $M{+}1$ sets of four-stage feature maps $\boldsymbol{f}_l^m{\in}\{\boldsymbol{f}_1^m,\boldsymbol{f}_2^m,\boldsymbol{f}_3^m,\boldsymbol{f}_4^m\}$, $m{\in}[1,M{+}1]$.
In the $l^{th}$ stage, the block number of each branch is $b_l{\in}\{4, 8, 16, 32\}$, the stride is $s_l{\in}\{4, 8, 16, 32\}$, and the channel dimension is $C_l{\in}\{64,128,320,512\}$.
Inside each stage, $M{+}1$ features are processed in the \emph{Hub2Fuse} paradigm:
At the \emph{hub} step, $M$ feature maps will be merged into one feature $\boldsymbol{f}^{q}$ via the proposed Self-Query Hub.
At the \emph{fusion} step, the merged feature $\boldsymbol{f}^{q}$ will be further fused with RGB feature by the cross-modal Feature Rectification Module (FRM)~\cite{liu2022cmx} and Feature Fusion Module (FFM)~\cite{liu2022cmx}, termed as $\boldsymbol{f}$.
These two modules enable better multimodal feature fusion and interaction, and are crucial when fusing RGB with sparse features, which will be shown in our experiments.
Between stages, $M{+}1$ feature maps will be restored via adding the fused feature $\boldsymbol{f}$, respectively. After the encoder, the {four-stage features} $\boldsymbol{f_l}{\in}\{\boldsymbol{f}_1,\boldsymbol{f}_2,\boldsymbol{f}_3,\boldsymbol{f}_4\}$ will be forwarded to the decoder for the segmentation prediction. We use the MLP decoder~\cite{xie2021segformer} as the segmentation head.

\subsection{Self-Query Hub}\label{sec:sq-hub}
To perform arbitrary-modal fusion, the Self-Query Hub~(SQ-Hub) is a crucial design to select the informative features of supplementary modalities before fusing with the RGB feature.
As shown in Fig.~\ref{fig:model}, given a set of $M$ supplementary features $\{\boldsymbol{f}^m|m{\in}[1,M], \boldsymbol{f}^m{\in}H{\times}W{\times}C\}$, a Self-Query module is applied to calculate the informative score mask $Q^m{\in}H{\times}W$ of each feature $\boldsymbol{f}^m$, as in Eq.~\eqref{eq:sq-hub1} and \eqref{eq:sq-hub2}. 
\begin{align}
\begin{split}\label{eq:sq-hub1}
    \boldsymbol{\hat{f}}^m &= \text{DW-Conv}_{3{\times}3}{(C, C)}(\boldsymbol{f}^m),
\end{split}\\
\begin{split}\label{eq:sq-hub2}
    Q^m &= \text{Sigmoid}(\text{Conv}(C, 1)(\boldsymbol{\hat{f}}^m)),
\end{split}
\end{align}
where the $\text{DW-Conv}_{3{\times}3}{(C_{in}, C_{out})}(\cdot)$ means a Depth-Wise convolution layer with a kernel size of $3{\times}3$.
After obtaining $M$ score masks through $M$ respective self-query modules, a cross-modal comparison is conducted between $M$ features $\{\boldsymbol{f}^m|m{\in}[1,M]\}$.
That is, each patch $p^q$ of the merged feature map $\boldsymbol{f}^q$ will be filled by the patch $p^m$ of $\{\boldsymbol{f}^m|m{\in}[1,M]\}$ with the highest score, \ie, the most effective patch among $M$ modalities.
It can be formalized as:
\begin{align}\label{eq:sq-hub3}
\begin{split}
    \boldsymbol{f}^q &=\{p^q|p^q{\in}{H{\times}W}\} \\
     &= \phi(\{\boldsymbol{f}^m{+}Q^m{\cdot}\boldsymbol{\hat{f}}^m|m{\in}[1,M]\}) \\
     &= \phi(\{p^m|p^m{\in}{H{\times}W},m{\in}[1,M]\}), 
\end{split}
\end{align}
where $\phi(\cdot)$ is an operation to select the maximum $p^m$ from $\{\boldsymbol{f}^m{+}Q^m{\cdot}\boldsymbol{\hat{f}}^m|m{\in}[1,M]\}$.
Then, the merged feature $\boldsymbol{f}^q$ is forwarded to the Parallel Pooling Mixer (PPX). 

\subsection{Parallel Pooling Mixer}\label{sec:ppx}
Another crucial design in CMNeXt is the Parallel Pooling Mixer (Fig.~\ref{fig:model}), which is proposed to efficiently and flexibly harvest discriminative cues from arbitrary-modal complements in the aforementioned SQ-Hub.
Given the merged feature map $\boldsymbol{f}^q{\in}H{\times}W{\times}C$ from SQ-Hub, a $7{\times}7$ DW-Conv layer is applied to aggregate local information.
The three parallel pooling layers are for capturing multi-scale modal features, which will be summed with the residual one and mixed by a $1{\times}1$ convolution.
Then, a Sigmoid function is used to calculate the attention for weighting. The first part of PPX can be written as:
\begin{align}\label{eq:ppx1}
\begin{split}
    \hat{\boldsymbol{f}^q}&=\text{DW-Conv}_{7{\times}7}(C, C)(\boldsymbol{f}^q), 
\end{split}\\
\begin{split}
    \hat{\boldsymbol{f}^q}&\coloneqq\sum_{k{\in}\{3,7,11\}}\text{Pool}_{k{\times}k}(\hat{\boldsymbol{f}^q}) + \hat{\boldsymbol{f}^q}, 
\end{split}\\
\begin{split}
    \boldsymbol{w}&=\text{Sigmoid}(\text{Conv}_{1{\times}1}(C, C)(\hat{\boldsymbol{f}^q})), 
\end{split}\\
\begin{split}
    \boldsymbol{f}^w&=\boldsymbol{w}{\cdot}\boldsymbol{f}^q+\boldsymbol{f}^q. 
\end{split}
\end{align}

Previous cross-modal fusion methods show that channel information is crucial~\cite{chen2020sagate,hu2019acnet}.
Inspired by this, we apply a Squeeze-and-Excitation (SE) module~\cite{hu2018senet} in the mixing part of PPX.
This structure is crucial since some channels of certain modalities do capture more significant information than others.
It can further engage more spatially-holistic knowledge in the channels of the cross-modal complements in SQ-Hub.
Thus, the weighted feature $\boldsymbol{f}^w$ is passed to a Feed-Forward Network (FFN) and a SE module~\cite{hu2018senet} for enhancing the channel information.
The second part of PPX can be written as:
\begin{equation}\label{eq:ppx2}
\begin{aligned}
    \hat{\boldsymbol{f}^w}&=\text{FFN}(C, C)(\boldsymbol{f}^w)+\text{SE}(\boldsymbol{f}^w). \\
\end{aligned}
\end{equation}
After the PPX block, $\hat{\boldsymbol{f}^w}$ is fused with RGB feature to form the final {fused feature} $\boldsymbol{f_l}{\in}\{\boldsymbol{f}_1,\boldsymbol{f}_2,\boldsymbol{f}_3,\boldsymbol{f}_4\}$ by using FRM\&FFM modules~\cite{liu2022cmx}, as shown in Fig.~\ref{fig:model}.

Compared with convolution-based MSCA~\cite{guo2022segnext}, pooling-based MetaFormer~\cite{yu2022metaformer}, fully-attentional FAN~\cite{zhou2022fan}, our PPX includes two advances: (1) parallel pooling layers for efficient weighting in the attention part; (2) channel-wise enhancement in the feature mixing part. Both characteristics of the PPX block help in highlighting the cross-modal fused feature spatial- and channel-wise, respectively. More comparisons will be presented in~\Cref{sec:experiments}.

\begin{figure*}[t]
    \centering
    \begin{minipage}{.5\textwidth}
        \begin{subfigure}[t]{.99\textwidth}
        \includegraphics[width=1.0\textwidth]{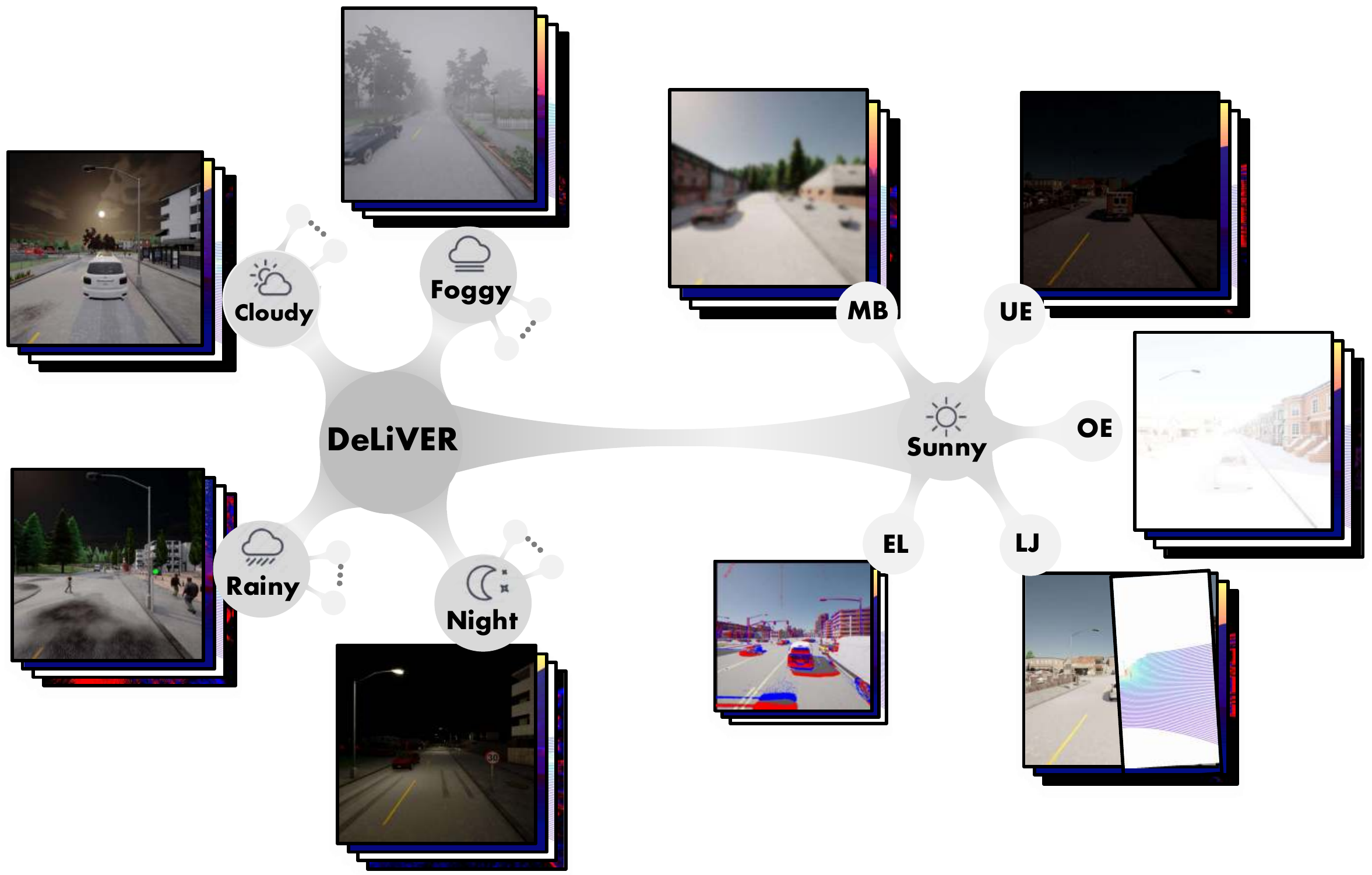}    
        \caption{\textbf{Structure and samples} of four adverse conditions and five failure cases.}
        \label{fig:dataset_structure}
        \end{subfigure}
    \end{minipage}    
    \begin{minipage}{.49\textwidth}
        \begin{subfigure}[t]{.99\textwidth}
            \input{Tables/deliver_stat}
        \end{subfigure} \\
        \begin{subfigure}[b]{.99\textwidth}
            \includegraphics[width=1.0\textwidth]{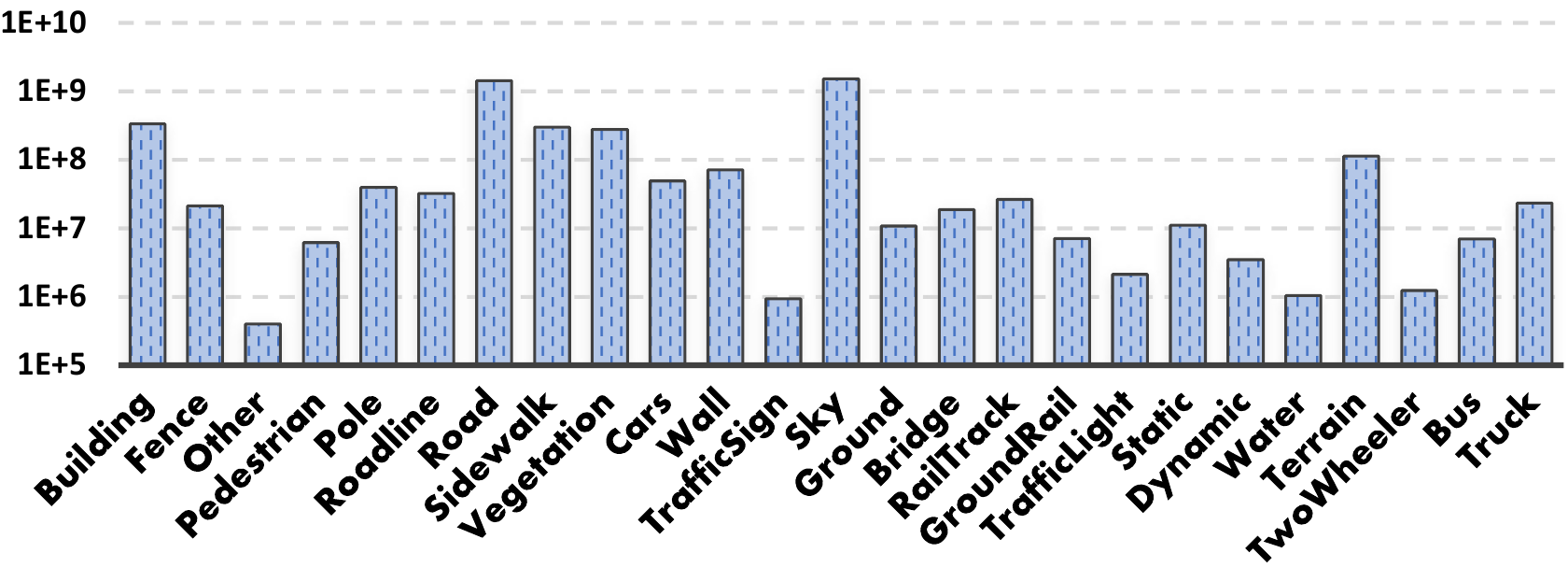}     
            \vskip -1ex
            \caption{\textbf{Distribution} of $25$ semantic classes in logarithmic scaling.}
            \label{fig:dataset_dist}
        \end{subfigure} 
    \end{minipage}%
    \caption{\textbf{\textsc{DeLiVER} multimodal dataset} including (a) four adverse conditions out of five conditions(\ie, \emph{cloudy}, \emph{foggy}, \emph{night-time}, \emph{rainy} and \emph{sunny}). Apart from normal cases, each condition has five corner cases (\ie, \textbf{MB}: Motion Blur; \textbf{OE}: Over-Exposure; \textbf{UE}: Under-Exposure; \textbf{LJ}: LiDAR-Jitter; and \textbf{EL}: Event Low-resolution). Each sample has six views. Each view has four modalities and two labels (\ie, semantic and instance). (b) is the data statistics. (c) is the data distribution of $25$ semantic classes.}
    \label{fig:dataset}
    \vskip -2ex
\end{figure*}

\section{The \textsc{\textbf{DeLiVER}} Multimodal Dataset}
\label{sec:dataset}

\noindent\textbf{Sensor settings and modalities.} 
As presented in Fig.~\ref{fig:dataset}, we spent the effort to create a large-scale multimodal segmentation dataset \textsc{\textbf{DeLiVER}} with \textbf{De}pth, \textbf{Li}DAR, \textbf{V}iews, \textbf{E}vent, \textbf{R}GB data, based on the CARLA simulator~\cite{dosovitskiy2017carla}. \textsc{DeLiVER} provides six mutually orthogonal views (\ie, \emph{front, rear, left, right, up, down}) of the same spatial viewpoint, \ie, a complete frame of data is encoded in the format of a panoramic cubemap.
The Field-of-View (FoV) of each view is $91^{\circ}{\times}91^{\circ}$ and the image resolution is $1042{\times}1042$.
All Depth, Views, and Event sensors use the same camera settings when the sensor is working properly.
According to the characteristics of recent LiDAR sensors~\cite{gao2021we_hungry}, we further customize a $64$ vertical channels virtual semantic LiDAR sensor to generate a point cloud of $1,728,000$ points per second with a FoV of $360^{\circ}{\times} (-30^{\circ}{\sim}10^{\circ})$ and a range of $100$ meters, so as to collect relatively dense LiDAR data.

\noindent\textbf{Adverse conditions and corner cases.} 
In addition to the multimodal setup, \textsc{DeLiVER} provides cases in two-fold, including four environmental conditions and five partial sensor failure cases~(Fig.~\ref{fig:dataset_structure}).
For environmental conditions, we consider \emph{cloudy}, \emph{foggy}, \emph{night}, and \emph{rainy} weather conditions other than only \emph{sunny} days.
The environmental conditions will cause variations in the position and illumination of the sun, atmospheric diffuse reflections, precipitation, and shading of the scene, introducing challenges for robust perception.
For sensor failure cases, we consider Motion Blur (MB), Over-Exposure (OE), and Under-Exposure (UE) common for RGB cameras.
LiDAR failures usually manifest as along-axis LiDAR-Jitter (LJ) due to fixation issues or rotational axis eccentricity, thus we add random angular jitters in the range of $[-1^\circ, 1^\circ]$ and position jitters of $[-1cm, 1cm]$ to the three axial directions of the LiDAR sensor.
Due to the circuit design, the resolution of the currently-used event sensors is limited~\cite{gallego2022event_survey}.
Thus, we customize an Event Low-resolution (EL) scenario with $0.25{\times}$ resolution for the event camera to simulate actual devices.

\noindent\textbf{Statistics and annotations.}
Including six views, \textsc{DeLiVER} has totally $47,310$ frames (Fig.~\ref{tab:deliver_stat}) with the size of $1042{\times}1042$. The $7,885$ front-view samples are divided into $3,983/2,005/1,897$ for training/validation/testing, respectively, each of which contains two types of annotations (\ie, semantic and instance segmentation labels).
Note that, we mainly discuss the front view and the semantic segmentation task in this work, while other views and instance segmentation will be future works.
To improve the class diversity of annotations~($25$ classes as in Fig.~\ref{fig:dataset_dist}), we modify and remap the semantic labels in the source code. Specifically, the \emph{Vehicles} class is subdivided into four fine-grained categories: \emph{Cars}, \emph{TwoWheeler}, \emph{Bus}, and \emph{Truck} for both the semantic camera and the semantic LiDAR, making \textsc{DeLiVER} compatible with popular segmentation datasets.

\section{Experiments}
\label{sec:experiments}

\subsection{Datasets and Implementation Details}
\noindent\textbf{KITTI-360}~\cite{liao2022kitti360} is a suburban driving dataset, having $49,004/12,276$ images at the size of $1408{\times}376$ for training/validation with $19$ classes.
To study RGB-Depth-Event-LiDAR fusion consistent with the \textsc{DeLiVER} dataset, we generate depth images and event data by using popular off-the-shelf models, \ie, AANet~\cite{xu2020aanet} and EventGAN~\cite{zhu2021eventgan}.

\noindent\textbf{MFNet}~\cite{ha2017mfnet} is an urban street dataset with $1,569$ RGB-Thermal pairs at the size of $640{\times}480$ with $8$ classes. $820$ pairs are collected during the day and the other $749$ are captured at night.
The training set consists of $50\%$ of the daytime- and $50\%$ of the nighttime images, whereas the validation- and test set respectively contains $25\%$ of the daytime- and $25\%$ of the nighttime images. 

\noindent\textbf{NYU Depth V2}~\cite{silberman2012nyuv2} is an indoor understanding dataset with $1,449$ RGB-Depth pairs at the size of $640{\times}480$, splitting into $795/654$ for training/testing with $40$ classes.

\noindent\textbf{UrbanLF}~\cite{sheng2022urbanlf} is a light field semantic segmentation dataset with both real-world and synthetic sets annotated in $14$ classes, respectively splitting into $580/80/164$ and $172/28/50$ samples for training/validation/testing. The real images have a size of $623{\times}432$, whereas the synthetic ones are of $640{\times}480$. Each sample is composed of $81$ sub-aperture images, leading to $81$ modalities.

\noindent\textbf{MCubeS}~\cite{liang2022mcubesnet} is a dataset with pairs of RGB, Near-Infrared (NIR), Degree of Linear Polarization (DoLP), and Angle of Linear Polarization (AoLP), to study semantic material segmentation of $20$ classes. It has $302/96/102$ image pairs for training/validation/testing at the size of $1224{\times}1024$.

\noindent\textbf{Implementation details.}
We train our models on four A100 GPUs with an initial learning rate (LR) of $6e^{-5}$, which is scheduled by the poly strategy with power $0.9$ over $200$ epochs.
The first $10$ epochs are to warm-up models with $0.1{\times}$ the original LR. We use cross-entropy loss function.
The optimizer is AdamW~\cite{loshchilov2017adamw} with epsilon $1e^{-8}$, weight decay $1e^{-2}$, and the batch size is $2$ on each GPU.
The images are augmented by random resize with ratio $0.5${--}$2.0$, random horizontal flipping, random color jitter, random gaussian blur, and random cropping to $1024{\times}1024$ on \textsc{DeLiVER}, while to their proposed sizes on other datasets.
To conduct comparisons, the ImageNet-1K~\cite{deng2009imagenet} pre-trained weight for the accompanying branch is not used on \textsc{DeLiVER} and KITTI-360, while the pre-trained weight for the RGB branch is applied on all datasets.

\subsection{Comparison against the State of the Art}
To verify the efficacy of our proposed CMNeXt framework, we conduct extensive experiments on six multimodal segmentation datasets.
The results and comparisons against the state-of-the-art are shown in Table~\ref{tab:res_all_six}.

\input{Tables/results_six_datasets}
\noindent\textbf{Results on \textsc{DeLiVER}.}
Table~\ref{tab:quad_modal} summarizes the extensive comparisons between our CMNeXt and other recent methods on \textsc{DeLiVER} dataset. 
Overall, CMNeXt sets the state of the art on the fusion of two to four modalities. 
While fusing RGB with Depth, Event, and LiDAR, the bi-modal CMNeXt yields sufficient improvements, compared to HRFuser~\cite{broedermann2022hrfuser} and TokenFusion~\cite{wang2022tokenfusion}.
This demonstrates the superiority of our \emph{Hub2Fuse} paradigm over the \emph{seperate} and \emph{joint} branch paradigm (Fig.~\ref{fig3:hrfuser} and Fig.~\ref{fig3:tokenfusion}), especially when fusing sparse modalities, \ie, Event and LiDAR.
From RGB-only to gradually fusing Depth, Events, and LiDAR, the mIoU scores of CMNeXt are gradually increased ($57.20\%{\rightarrow}63.58\%{\rightarrow}64.44\%{\rightarrow}66.30\%$), showing the advance of arbitrary-modal fusion for segmentation.
Thanks to the complementary features from other modalities, our quad-modal CMNeXt outperforms the RGB-only baseline SegFormer~\cite{xie2021segformer} by a significant margin of ${+}9.10\%$.

\noindent\textbf{Results on KITTI-360.}
In Table~\ref{tab:quad_modal}, apart from the \textsc{DeLiVER} dataset with adverse cases, we further conduct equivalent experiments on KITTI-360~\cite{liao2022kitti360} which only contains normal scenes.
We found that most of the multimodal fusion methods on KITTI-360 did not bring the expected high improvement.
There are two conjectures: The samples are collected in suburbs and are composed of video sequences, resulting in insufficient scene diversity; The depth- and event data are generated from RGB sequences, resulting in limited modal differences.
Thus, the segmentation output relies on the RGB segmentation, and adding modalities might be redundant.
Nonetheless, our quad-modal CMNeXt achieves a ${+}0.80\%$ gain compared to the RGB-only baseline~\cite{xie2021segformer}. Besides, our bi-modal CMNeXt performs superior to CMX~\cite{liu2022cmx} by ${+}1.56\%$ to ${+}2.85\%$. 
When fusing three to four modalities, CMNeXt has respective ${+}17.52\%$, ${+}13.94\%$, and ${+}15.08\%$ gains compared to HRFuser~\cite{broedermann2022hrfuser}.

\noindent\textbf{RGB-T and RGB-D segmentation.}
As shown in Table~\ref{tab:MFNet} and \ref{tab:NYU}, we further conduct experiments on bi-modal datasets, MFNet~\cite{ha2017mfnet} and NYU Depth V2~\cite{silberman2012nyuv2}, which comprise dense thermal and depth data as supplementary information.
Our CMNeXt achieves the state of the art on both datasets.
Using MiT-B4~\cite{xie2021segformer}, CMNeXt outperforms CMX with ${+}0.2\%$ on MFNet. Besides, on the NYU Depth V2 dataset, it is comparable to CMX with MiT-B5.
It proves the benefits of our PPX block in CMNeXt over the Multi-Head Self-Attention (MHSA) block used by CMX. 

\noindent\textbf{Light field semantic segmentation.}
Towards arbitrary-modal fusion for semantic segmentation, we apply CMNeXt on the UrbanLF dataset~\cite{sheng2022urbanlf}, in which each sample is composed of $81$ sub-aperture light field modalities. As shown in Table~\ref{tab:urbanlf}, CMNeXt surpasses the previous state of the art, OCR-LF~\cite{sheng2022urbanlf}, in both real-world and synthetic scenes, even with fewer modalities ($33{\rightarrow}8$).
Due to the similarity between modalities in this dataset, it is challenging to extract diverse complementary features. Nonetheless, by fusing up to $80$ light field images, CMNeXt reaches respective $83.11\%$ and $81.02\%$ in mIoU on real and synthetic sets.

\noindent\textbf{Multimodal material segmentation.}
To verify multimodal fusion in material recognition, we conduct experiments on the MCubeS dataset~\cite{liang2022mcubesnet} which also contains four modalities.
As shown in Table~\ref{tab:MCubeS}, our quad-modal CMNeXt exceeds other quad-modal models and attains the top performance of $51.54\%$, with a significant increase $8.68\%$ over MCubeSNet~\cite{liang2022mcubesnet}.
In addition, CMNeXt has incremental improvements when gradually adding AoLP, DoLP, and NIR modalities.
The results on multimodal material segmentation are consistent with the ones of arbitrary-modal segmentation on our \textsc{DeLiVER} dataset.

\subsection{Ablation Studies}
\input{Tables/result_condition}
\noindent\textbf{Analysis in adverse weather conditions.} In Table~\ref{tab:res_condition}, we compare CMNeXt against mainstream multimodal fusion paradigms in different conditions including adverse weather- and partial sensor failure scenarios.
It can be seen that despite being efficient, TokenFusion~\cite{wang2022tokenfusion} suffers in these conditions as effective information is discarded in their token replacement.
Due to the proposed SQ-Hub for selecting effective features, CMNeXt significantly improves the performance compared to the previous CMX~\cite{liu2022cmx} and HRFuser~\cite{broedermann2022hrfuser}.
When fusing more modalities, HRFuser tends to induce much more overhead (${+}6.00$ GFLOPs when adding a branch), whereas CMNeXt brings great mIoU gains at only slight computation increase (${<}1.30$ GFLOPs).
Compared with the RGB baseline, the full RGB-D-E-L CMNeXt overall improves the accuracy by $9.10\%$ on average for different conditions, in particular for the nighttime (${+}12.01\%$) and the rainy (${+}8.81\%$) scenarios.

\noindent\textbf{Analysis in sensor failure cases.}
In the Event Low-resolution (\textbf{EL}) case of Table~\ref{tab:res_condition}, from the fusion of RGB-D to RGB-D-E, the accuracy of HRFuser~\cite{broedermann2022hrfuser} is degraded, however, the one of CMNeXt is improved ($63.35\%{\rightarrow}66.11\%$).
This is also observed in the case of LiDAR Jitter (\textbf{LJ}), where the performance of CMNeXt is increased ($65.37\%{\rightarrow}65.92\%$) by fusing from D-E to D-E-L.
These results demonstrate the ability of CMNeXt to combat sensor failures, thanks to SQ-Hub for selecting informative features.
Compared to the RGB baseline, CMNeXt obtains a ${+}22.56\%$ gain in the Under-Exposure (\textbf{UE}) case.

\input{Tables/ablation}
\noindent\textbf{Ablation of the CMNeXt architecture.}
As shown in Table~\ref{tab:ablation}, we ablate our CMNeXt architecture.
When removing the addition operation of supplementary modalities, the performance slightly decreases.
Without the SQ-Hub for dynamically harvesting complementary cues, the supplementary modalities are directly added and the mIoU declines by $1.89\%$.
When using the MSCA from SegNeXt~\cite{guo2022segnext} instead of our PPX, the accuracy clearly drops.
Ablating the SE block in PPX for channel processing incurs a mIoU downgrade of $3.03\%$, which indicates that the spatially-holistic knowledge in channels contribute a lot to the multimodal fusion.
The FRM\&FFM modules also play important roles in facilitating comprehensive cross-modal interactions between the RGB representation and the supplementary representation extracted via SQ-Hub.
The results verify that the hub and fusion steps in our proposed \emph{Hub2Fuse} paradigm are fundamental to arbitrary multimodal segmentation.

\input{Tables/compare_attn}
\noindent\textbf{Comparison of token mixing blocks.}
As shown in Table~\ref{tab:compare_attn}, we first compare PPX against convolutional-, attentional, and pooling-based blocks when ported on our CMNeXt architecture as the accompanying branch for supplementary-modal features.
PPX achieves the best mIoU score, while remaining highly efficient with few parameters.
While the PoolFormer~\cite{yu2022metaformer} has less parameters and GLFOPs, it is also less effective for harvesting cross-modal cues.
PPX surpasses the MHSA in SegFormer~\cite{xie2021segformer}, ConvNeXt~\cite{liu2022convnext}, the fully attentional block in FAN~\cite{zhou2022fan}, the $g^n$Conv in HorNet~\cite{rao2022hornet}, the MSCA in SegNeXt~\cite{guo2022segnext}.
Compared with the P2T block~\cite{wu2022p2t} adapting pyramid pooling in self-attention, our PPX is both more efficient and accurate, making it ideally suitable for learning complementary features towards arbitrary multimodal fusion. 

After confirming that PPX block in the accompanying branch, for the RGB branch, we follow CMX~\cite{liu2022cmx} and use MHSA blocks from SegFormer. In spite of moderate complexity, MSHA~\cite{xie2021segformer}+PPX achieves higher accuracy than PPX+PPX and MSCA~\cite{guo2022segnext}+PPX, indicating that self-attention excels at learning from the dense RGB representation in multimodal semantic segmentation.

\begin{figure}
    \centering
    \includegraphics[width=1\columnwidth]{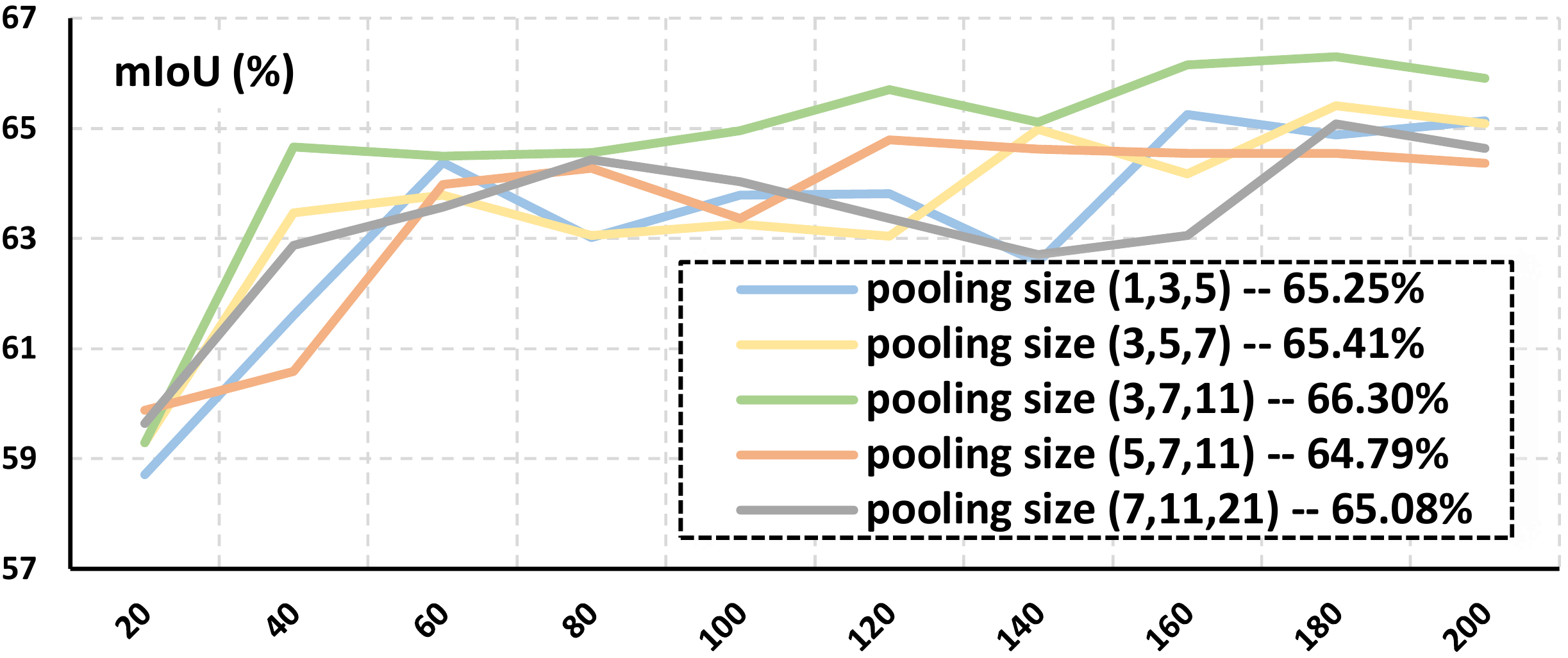}
    \caption{Training curves of different pooling sizes in PPX.}
    \label{fig:pooling_size}    
\end{figure}
\noindent\textbf{Parameter study on the pooling sizes.}
In Fig.~\ref{fig:pooling_size}, we investigate a variety of pooling sizes in PPX on our \textsc{DeLiVER} dataset, confirming the set of $\{3,7,11\}$ yields the best mIoU.

\begin{figure}
    \centering
    \includegraphics[width=1\columnwidth]{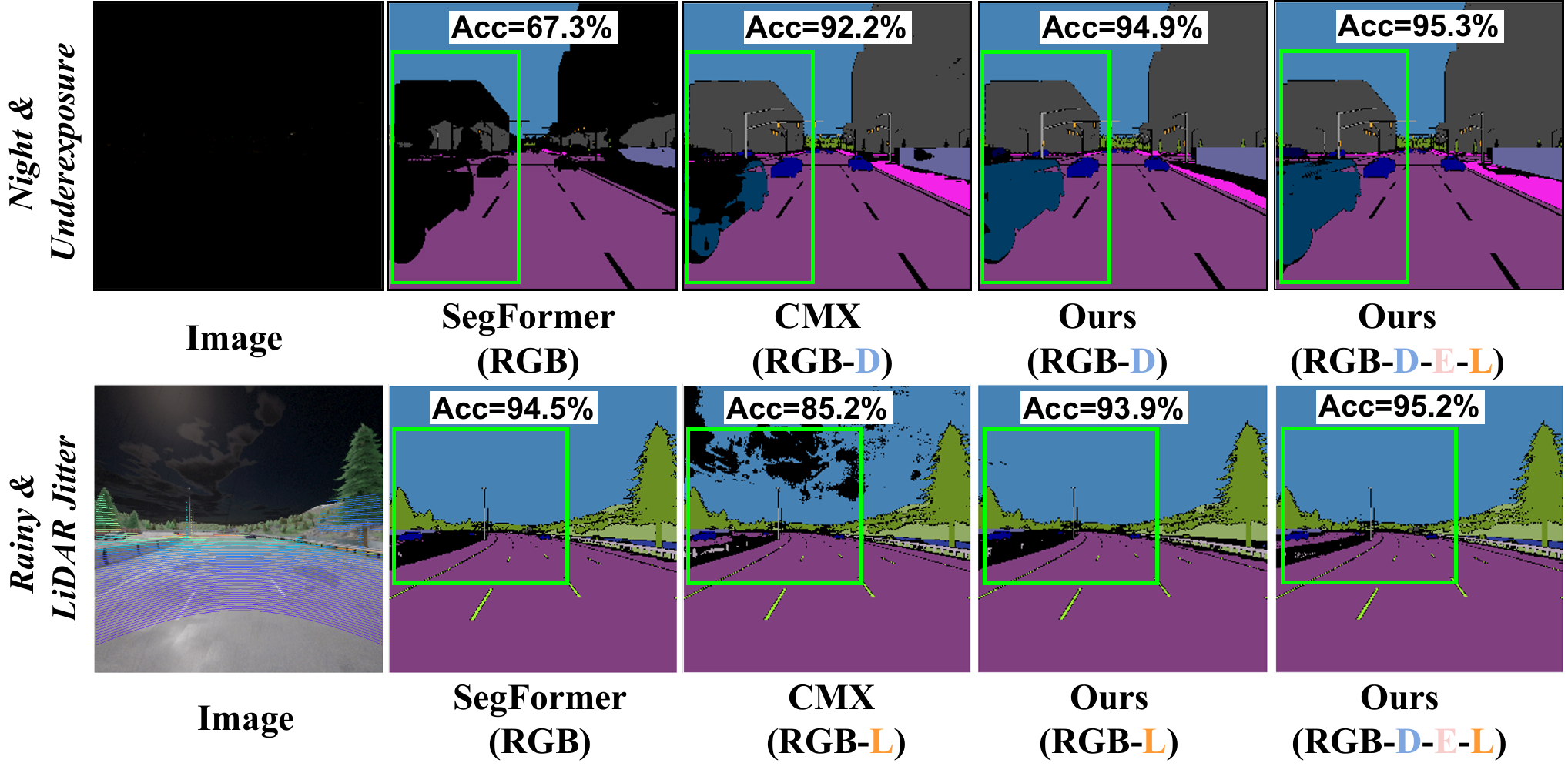}
    \caption{Visualization of segmentation results.}
    \label{fig:qualitative_vis}
\end{figure}
\noindent\textbf{Visualization of arbitrary-modal segmentation.}
In Fig.~\ref{fig:qualitative_vis}, we show semantic segmentation results of our CMNeXt against the RGB-only SegFormer~\cite{xie2021segformer} and the RGB-X CMX~\cite{liu2022cmx}.
It can be seen that in the dark night with under-exposure, the RGB-only SegFormer hardly segments the close vehicle, while the RGB-D CMNeXt clearly outperforms CMX.
Our RGB-D-E-L CMNeXt further enhances the performance and yields more complete segmentation.
In the partial sensor failure scenario with LiDAR jitter, CMX produces unsatisfactory rainy scene parsing results. 
Our RGB-LiDAR model is barely affected by the sensing data mis-alignment and the quad-modal CMNeXt further robustifies the full scene segmentation.


\section{Conclusion}
\input{Tex_content/conclusion}


\clearpage
{\small
\bibliographystyle{ieee_fullname}
\bibliography{main}
}

\input{Tex_content/appendix}

\end{document}

%% file: Tex_content/abstract.tex
Multimodal fusion can make semantic segmentation more robust. However, fusing an arbitrary number of modalities remains underexplored. To delve into this problem, we create the \textsc{DeLiVER} arbitrary-modal segmentation benchmark, covering \underline{De}pth, \underline{Li}DAR, multiple \underline{V}iews, \underline{E}vents, and \underline{R}GB. Aside from this, we provide this dataset in four severe weather conditions as well as five sensor failure cases to exploit modal complementarity and resolve partial outages. To make this possible, we present the arbitrary cross-modal segmentation model \textsc{CMNeXt}. It encompasses a \emph{Self-Query Hub~(SQ-Hub)} designed to extract effective information from any modality for subsequent fusion with the RGB representation and adds only negligible amounts of parameters (${\sim}0.01M$) per additional modality. On top, to efficiently and flexibly harvest discriminative cues from the auxiliary modalities, we introduce the simple \emph{Parallel Pooling Mixer (PPX)}. With extensive experiments on a total of six benchmarks, our \textsc{CMNeXt} achieves state-of-the-art performance on the \textsc{DeLiVER}, KITTI-360, MFNet, NYU Depth V2, UrbanLF, and MCubeS datasets, allowing to scale from $1$ to $81$ modalities. On the freshly collected \textsc{DeLiVER}, the quad-modal \textsc{CMNeXt} reaches up to $66.30\%$ in mIoU with a ${+}9.10\%$ gain as compared to the mono-modal baseline.\footnote{The \textsc{DeLiVER} dataset and our code will be made publicly available at: \url{https://jamycheung.github.io/DELIVER.html}.}

%% file: Tex_content/related_work.tex
\noindent\textbf{Semantic segmentation} has experienced striking progress since fully convolutional networks~\cite{long2015fcn} introducing the end-to-end per-pixel classification paradigm, which was enhanced by capturing multi-scale features~\cite{chen2017deeplab,chen2018deeplabv3+,hou2020strip,zhao2017pspnet}, appending channel- and self-attention blocks~\cite{choi2020hanet,fu2019danet,huang2019ccnet,yuan2021ocnet}, refining context priors~\cite{jin2021mining,lin2017refinenet,yu2020context_prior,zhang2018context_encoding}, and leveraging edge cues~\cite{borse2021inverseform,ding2019boundary_propagation,li2020improving_decoupled_body_edge,takikawa2019gated}.
Recently, with the application of vision transformers in recognition tasks, dense prediction transformers~\cite{dong2022cswin,lee2022mpvit,wang2021pvt,yuan2021hrformer} and semantic segmentation transformers~\cite{gu2022hrvit,strudel2021segmenter,zhang2022segvit,zheng2021setr} emerge, along with the mask classification paradigm~\cite{cheng2021maskformer,cheng2022mask2former} to jointly handle things and stuff segmentation.
Following the general architecture of transformers, attention-based token mixing has been substituted with MLP-based~\cite{chen2021cyclemlp,hou2022vision_permutator,lian2021asmlp}, pooling~\cite{yu2022metaformer}, and convolutional~\cite{guo2022segnext,guo2022visual} blocks.
While these works achieve great improvements on mainstream image segmentation benchmarks, they still suffer under real-world conditions where RGB images do not offer sufficient textures like low-illumination and fast-moving scenarios.

\noindent\textbf{Multimodal semantic segmentation} has been considered by harvesting complementary features from supplementary modalities such as depth~\cite{cao2021shapeconv,chen2021sgnet,ying2022uctnet,zhou2020rgb_coattention}, thermal~\cite{shivakumar2020pst900,wu2022complementarity,zhang2021abmdrnet}, polarization~\cite{kalra2020deep_polarization,mei2022pgsnet,xiang2021polarization}, events~\cite{alonso2019ev_segnet,zhang2021issafe}, LiDAR~\cite{yan20222dpass,zhuang2021pmf}, and optical flow~\cite{rashed2019optical_flow}.
To scale from modality-specific fusion to unified fusion, CMX~\cite{liu2022cmx} tackles RGB-X segmentation with multi-level cross-modal interactions, whereas channel- and token exchanges are explored in~\cite{wang2022tokenfusion,wang2020deep_channel_exchanging,wang2020learning_asymmetric_fusion}.
Additional multimodal fusion methods address object detection~\cite{li2022deepfusion,song2021exploiting}, medical and material segmentation~\cite{liang2022mcubesnet,xing2022nestedformer}, as well as flow estimation~\cite{liu2022camliflow}.
Most of these works focus on fusing complementary cues, but they do not fully consider multimodal learning in scenarios where some modalities fail.
To this end, we propose CMNeXt, a universal multimodal semantic segmentation framework with arbitrary-modal complements.
Unlike previous modality-specific fusion methods~\cite{hu2019acnet,mei2022pgsnet,zhang2021issafe}, CMNeXt scales from bi-modal scenarios like RGB-D parsing to arbitrary-modal fusion like light field segmentation with virtually $81$ modalities.
In addition, we provide a \textsc{DeLiVER} benchmark to foster multimodal learning.
While there are some existing datasets~\cite{gehrig2021eventscape,sekkat2022synwoodscape,testolina2022selma} based on the CARLA simulator~\cite{dosovitskiy2017carla}, our dataset not only provides diverse sensing data but also sensor-failure cases for robust semantic understanding.

%% file: Tables/deliver_stat.tex
\caption{\textbf{Statistic} of different data splits and views.}
\label{tab:deliver_stat}
\resizebox{\columnwidth}{!}{
\setlength{\tabcolsep}{1mm}{
\begin{tabular}{l|rrrrr|rr|r} 
\toprule
\textbf{Split} & \textbf{Cloudy} & \textbf{Foggy} & \textbf{Night} & \textbf{Rainy} & \textbf{Sunny}& \textbf{Normal} & \textbf{Corner}  & \textbf{Total}  \\\midrule\hline
Train          & 794    & 795   & 797   & 799    & 798   & 2585   & 1398   & 3983   \\
Val            & 398    & 400   & 410   & 398    & 399   & 1298   & 707    & 2005   \\
Test           & 379    & 379   & 379   & 380    & 380   & 1198   & 699    & 1897   \\\hline
Front-view     & 1571   & 1574  & 1586  & 1577   & 1577  & 5081   & 2804   & 7885   \\\hline
All six views  & 9426   & 9444  & 9516  & 9462   & 9462  & 30486  & 16824  & 47310  \\
\bottomrule
\end{tabular}
}
}

%% file: Tables/results_six_datasets.tex
\begin{table*}[!t]
\centering
\caption{\textbf{Results on six multimodal semantic segmentation datasets}. The KITTI-360~\cite{liao2022kitti360} and our DeLiVER datasets have up to four modalities. The MFNet~\cite{ha2017mfnet} and NYU Depth V2~\cite{silberman2012nyuv2} datasets are dual-modal with respective RGB-Thermal and RGB-Depth modalities. The UrbanLF~\cite{sheng2022urbanlf} has up to 81 sub-aperture light-filed images. The quad-modal MCubeS dataset~\cite{liang2022mcubesnet} is for material segmentation.}
\vskip -1ex
\label{tab:res_all_six}
    \begin{subtable}[ht]{1.1\columnwidth}
    \caption{Results on KITTI-360 and \textsc{DeLiVER} datasets.}
    \label{tab:quad_modal}
    \setlength{\tabcolsep}{2pt}
    \resizebox{\columnwidth}{!}{    
    \renewcommand{\arraystretch}{1.22}
	\begin{tabular}{l|cc|c:c}
    \toprule
    \textbf{Method}& \textbf{Modal} & \textbf{Backbone} & \textbf{KITTI-360} & \textbf{DeLiVER} \\
    \midrule\hline
    HRFuser~\cite{broedermann2022hrfuser} & RGB & HRFormer-T  & 53.20 & 47.95 \\
    SegFormer~\cite{xie2021segformer} & RGB & MiT-B2 & 67.04 &  57.20 \\
    \hline
    HRFuser~\cite{broedermann2022hrfuser} & RGB-Depth & HRFormer-T &49.32 & 51.88\\
    TokenFusion~\cite{wang2022tokenfusion} & RGB-Depth & MiT-B2&  57.44 & 60.25\\
    CMX~\cite{liu2022cmx} & RGB-Depth & MiT-B2& 64.43 & 62.67\\
    \rowcolor{gray!15} CMNeXt & RGB-Depth & MiT-B2 & 65.09& 63.58 \\
    \hline
    HRFuser~\cite{broedermann2022hrfuser} & RGB-Event & HRFormer-T&44.85 & 42.22\\
    TokenFusion~\cite{wang2022tokenfusion} & RGB-Event & MiT-B2 & 55.97& 45.63 \\
    CMX~\cite{liu2022cmx} & RGB-Event & MiT-B2 & 64.03& 56.52\\
    \rowcolor{gray!15} CMNeXt & RGB-Event & MiT-B2 &  66.13& 57.48 \\
    \hline
    HRFuser~\cite{broedermann2022hrfuser} & RGB-LiDAR & HRFormer-T &48.74& 43.13\\
    TokenFusion~\cite{wang2022tokenfusion} & RGB-LiDAR & MiT-B2 & 54.55& 53.01\\
    CMX~\cite{liu2022cmx} & RGB-LiDAR & MiT-B2 & 64.31& 56.37\\
    \rowcolor{gray!15} CMNeXt & RGB-LiDAR & MiT-B2 &  65.26& 58.04 \\
    \hline
    HRFuser~\cite{broedermann2022hrfuser}& RGB-D-Event & HRFormer-T & 50.21& 51.83 \\
    \rowcolor{gray!15} CMNeXt & RGB-D-Event & MiT-B2 &  67.73& 64.44 \\
    \hline
    HRFuser~\cite{broedermann2022hrfuser}& RGB-D-LiDAR & HRFormer-T&52.61 & 52.72 \\
    \rowcolor{gray!15} CMNeXt & RGB-D-LiDAR & MiT-B2& 66.55 & 65.50 \\
    \hline
    HRFuser~\cite{broedermann2022hrfuser}& RGB-D-E-Li & HRFormer-T & 52.76& 52.97\\
    \rowcolor{gray!15} CMNeXt & RGB-D-E-Li & MiT-B2 & \textbf{67.84} & \textbf{66.30} \\
    \bottomrule
    \end{tabular}
    }
    \end{subtable}%
    \begin{subtable}[ht]{\columnwidth}
    \begin{subtable}[ht]{0.55\columnwidth}
    \centering
    \caption{Results on MFNet.}
    \label{tab:MFNet}
    \resizebox{\columnwidth}{!}{
    \renewcommand{\arraystretch}{1}
    \setlength{\tabcolsep}{1mm}{
    \begin{tabular}{l|c|c}
    \toprule
    \textbf{Method} & \textbf{Modal} & \textbf{mIoU}\\
    \midrule\hline
    SwinT~\cite{liu2021swin} & RGB & 49.0 \\
    SegFormer~\cite{xie2021segformer} & RGB & 52.0\\
    \hline
    ACNet~\cite{hu2019acnet} & RGB-T & 46.3 \\
    FuseSeg~\cite{sun2021fuseseg} & RGB-T & 54.5 \\
    ABMDRNet~\cite{zhang2021abmdrnet} & RGB-T & 54.8 \\
    LASNet~\cite{li2022lasnet} & RGB-T & 54.9 \\
    FEANet~\cite{deng2021feanet} & RGB-T & 55.3 \\
    MFTNet~\cite{zhou2022mftnet} & RGB-T & 57.3 \\
    GMNet~\cite{zhou2021gmnet} & RGB-T & 57.3 \\
    DooDLeNet~\cite{frigo2022doodlenet} & RGB-T & 57.3 \\
    CMX (MiT-B2)~\cite{liu2022cmx} & RGB-T & 58.2\\
    CMX (MiT-B4)~\cite{liu2022cmx} & RGB-T & 59.7\\
    \rowcolor{gray!15} CMNeXt (MiT-B4) & RGB-T & \textbf{59.9}\\
    \bottomrule
    \end{tabular}
    }
    }
    \end{subtable}%
    \hspace{\fill}
    \begin{subtable}[ht]{0.44\columnwidth}
    \centering
    \caption{Results on NYU Depth V2.}
    \label{tab:NYU}
    \resizebox{\columnwidth}{!}{
    \renewcommand{\arraystretch}{1}
    \setlength{\tabcolsep}{1mm}{
    \begin{tabular}{l|c}
    \toprule
    \textbf{Method} & \textbf{mIoU}  \\
    \midrule\hline
    ACNet~\cite{hu2019acnet} & 48.3 \\
    SGNet~\cite{chen2021sgnet}  & 51.1 \\
    ShapeConv~\cite{cao2021shapeconv} & 51.3  \\
    NANet~\cite{zhang2021nanet} & 52.3 \\
    SA-Gate~\cite{chen2020sagate} & 52.4  \\
    PGDENet~\cite{zhou2022pgdenet} & 53.7 \\
    \hline
    TokenFusion~\cite{wang2022tokenfusion} & 54.2 \\
    TransD-Fusion~\cite{wu2022transdfusion} & 55.5 \\
    MultiMAE~\cite{bachmann2022multimae} & 56.0 \\
    Omnivore~\cite{girdhar2022omnivore} & 56.8 \\
    CMX (MiT-B4)~\cite{liu2022cmx} & 56.3  \\
    CMX (MiT-B5)~\cite{liu2022cmx} & \textbf{56.9}  \\
    \rowcolor{gray!15} CMNeXt (MiT-B4) & \textbf{56.9} \\
    \bottomrule
    \end{tabular}
    }
    }
    \end{subtable}
    \begin{subtable}[ht]{0.54\columnwidth}
    \caption{Results on UrbanLF-Real and -Syn.}
    \label{tab:urbanlf}
    \resizebox{\columnwidth}{!}{
    \renewcommand{\arraystretch}{1.1}
    \setlength{\tabcolsep}{1mm}{
    \begin{tabular}{@{}l|c|cc@{}}
    \toprule
    \textbf{Method}& \textbf{Modal} & \textbf{Real} & \textbf{Syn} \\
    \midrule\hline
    PSPNet~\cite{zhao2017pspnet} & RGB  & 76.34 & 75.78\\
    OCR~\cite{yuan2020ocr} & RGB  & 78.60 & 79.36 \\
    SegFormer~\cite{xie2021segformer} (B4) & RGB & 82.20 & 78.53 \\
    \hline
    DAVSS~\cite{zhuang2020video} & Video &   75.91 & 74.27 \\
    TMANet~\cite{wang2021temporal} & Video & 77.14 & 76.41 \\
    \hline
    ESANet~\cite{seichter2021esanet}& RGB-D & \textit{n.a.} & 79.43\\
    SA-Gate~\cite{chen2020sagate} & RGB-D &  \textit{n.a.} & 79.53\\
    \hline
    PSPNet-LF~\cite{sheng2022urbanlf} & RGB-LF33 &  78.10 & 77.88\\
    OCR-LF~\cite{sheng2022urbanlf} & RGB-LF33 & 79.32& 80.43\\
    \rowcolor{gray!15} CMNeXt (MiT-B4) & RGB-LF8 & \textbf{83.22} & 80.74 \\
    \rowcolor{gray!15} CMNeXt (MiT-B4) & RGB-LF33 & 82.62 & 80.98 \\
    \rowcolor{gray!15} CMNeXt (MiT-B4) & RGB-LF80 & {83.11} & \textbf{81.02} \\
    \bottomrule
    \end{tabular}
    }}
    \end{subtable}%
    \hspace{\fill}
    \begin{subtable}[ht]{0.45\columnwidth}
    \caption{Results on MCubeS.}
    \label{tab:MCubeS}
    \resizebox{\columnwidth}{!}{
    \renewcommand{\arraystretch}{1.1}
    \setlength{\tabcolsep}{1mm}{
    \begin{tabular}{@{}l|c|c@{}}
    \toprule
    \textbf{Method}& \textbf{Modal} & \textbf{mIoU} \\
    \midrule\hline
    DRConv~\cite{chen2021drconv} &RGB-A-D-N & 34.63 \\
    DDF~\cite{zhou2021ddf} &RGB-A-D-N & 36.16 \\
    TransFuser~\cite{prakash2021transfuser} &RGB-A-D-N & 37.66 \\
    MMTM~\cite{joze2020mmtm} &RGB-A-D-N & 39.71 \\
    FuseNet~\cite{hazirbas2016fusenet} &RGB-A-D-N & 40.58 \\\hline
    MCubeSNet~\cite{liang2022mcubesnet} &RGB & 33.70 \\
    \rowcolor{gray!15} CMNeXt (MiT-B2) &RGB & 48.16 \\ 
    MCubeSNet~\cite{liang2022mcubesnet} &RGB-A & 39.10 \\
    \rowcolor{gray!15} CMNeXt (MiT-B2) &RGB-A & 48.42 \\ 
    MCubeSNet~\cite{liang2022mcubesnet} &RGB-A-D & 42.00 \\
    \rowcolor{gray!15} CMNeXt (MiT-B2) &RGB-A-D & 49.48 \\ 
    MCubeSNet~\cite{liang2022mcubesnet} &RGB-A-D-N & 42.86 \\
    \rowcolor{gray!15} CMNeXt (MiT-B2) &RGB-A-D-N & \textbf{51.54} \\ 
    \bottomrule
    \end{tabular}  
    }}
    \end{subtable}
    \end{subtable}
\vskip -1ex
\end{table*}

%% file: Tables/result_condition.tex
\begin{table*}
\centering
\caption{\textbf{Results on adverse conditions of \textsc{DeLiVER}}. Sensor failure cases are \textbf{MB}: Motion Blur; \textbf{OE}: Over-Exposure; \textbf{UE}: Under-Exposure; \textbf{LJ}: LiDAR-Jitter; and \textbf{EL}: Event Low-resolution. The number of parameters (\#Params) and GFLOPs are counted in~$512{\times}512$.}
\label{tab:res_condition}
\resizebox{\textwidth}{!}{
\renewcommand{\arraystretch}{1.}
\setlength{\tabcolsep}{1mm}{
\begin{tabular}{lll|ccccc|ccccc|c} 
\toprule
\textbf{Model-modality} & \textbf{\footnotesize{\#Params(M)}} & \textbf{\footnotesize{GFLOPs}} & \textbf{Cloudy} & \textbf{Foggy} & \textbf{Night} & \textbf{Rainy} & \textbf{Sunny} & \textbf{MB} & \textbf{OE} & \textbf{UE} & \textbf{LJ} & \textbf{EL} & \textbf{Mean}  \\
\midrule\hline
HRFuser-RGB       & 29.89       & 217.5 & 49.26  & 48.64 & 42.57 & 50.61  & 50.47 & 48.33 & 35.13 & 26.86 & 49.06 & 49.88 & 47.95  \\
SegFormer-RGB        & 25.79       & 38.93  & 59.99  & 57.30 & 50.45 & 58.69  & 60.21 & 57.28 & 56.64 & 37.44 & 57.17 & 59.12 & 57.20  \\ \hline
TokenFusion-RGB-D &26.01&54.96&50.92&52.02&43.37&50.70&52.21&49.22&46.22&36.39&49.58&49.17&49.86      \\
CMX-RGB-D         & 66.57       & 65.68  & 63.70  & 62.77 & 60.74 & 62.37  & 63.14 & 59.50 & 60.14 & 55.84 & 62.65 & 63.26 & 62.66  \\ \hline
HRFuser-RGB-D     & 30.46       & 223.0 & 54.80  & 51.48 & 49.51 & 51.55  & 52.12 & 50.92 & 41.51 & 44.00 & 54.10 & 52.52 & 51.88  \\ 
HRFuser-RGB-D-E   & 31.04~\obf{+0.57}       & 229.0~\obf{+6.00} & 54.04  & 50.83 & 50.88 & 51.13  & 52.61 & 49.32 & 41.75 & 47.89 & 54.65 & 52.33 & 51.83  \\
HRFuser-RGB-D-E-L & 31.61~\obf{+0.57}       & 235.0~\obf{+6.00} & 56.20  & 52.39 & 49.85 & 52.53  & 54.02 & 49.44 & 46.31 & 46.92 & 53.94 & 52.72 & 52.97  \\ \hline
CMNeXt-RGB-D      & 58.69       & 62.94  & 67.21  & 62.79 & 61.64 & 62.95  & 65.26 & 61.00 & 64.64 & 58.71 & 64.32 & 63.35 & 63.58  \\ 
CMNeXt-RGB-D-E    & 58.72~\obf{+0.03}       & 64.19~\obf{+1.25}  & 68.28 & 63.28 & 62.64 & 63.01 & 66.06 & 62.58 & 64.44 & 58.73 & 65.37 & 65.80 & 64.44  \\
CMNeXt-RGB-D-E-L  & 58.73~\obf{+0.01}       & 65.42~\obf{+1.23}  & 68.70 & 65.67 & 62.46 & 67.50 & 66.57 & 62.91 & 64.59 & 60.00 & 65.92 & 65.48 & 66.30  \\
\textit{w.r.t.} SegFormer-RGB   &   &  & ~\gbf{+8.71}   & ~\gbf{+8.37}  & ~\gbf{+12.01} & ~\gbf{+8.81}   & ~\gbf{+6.36}  & ~\gbf{+5.63}       & ~\gbf{+7.95}         & ~\gbf{+22.56}         & ~\gbf{+8.75}        & ~\gbf{+6.36}        & ~\gbf{+9.10}   \\
\bottomrule
\end{tabular}
}
}
\end{table*}

%% file: Tables/ablation.tex
\begin{table}
\centering
\caption{\textbf{Ablation study} of the CMNeXt architecture.}    
\label{tab:ablation}
\resizebox{\columnwidth}{!}{
\setlength{\tabcolsep}{1mm}{
\begin{tabular}{l@{}cc|l} 
\toprule
\textbf{Structure}       & \textbf{\#Params(M)} & \textbf{GFLOPs} & \textbf{mIoU(\%)}   \\\midrule\midrule
CMNeXt          & 58.73       & 65.42  & 66.30  \\ \hdashline
\quad -- without Addition    & 58.73       & 65.42  & 64.56~\obf{-1.74}  \\
\quad -- without SQ-Hub  & 58.70       & 65.36  & 64.41~\obf{-1.89}  \\
\quad -- with MSCA instead PPX     & 61.95       & 68.42  & 63.94~\obf{-2.36}  \\
\quad -- without SE in PPX      & 58.73       & 65.41  & 63.27~\obf{-3.03}  \\
\quad -- without FRM     & 48.71       & 64.79  & 62.71~\obf{-3.59}  \\
\quad -- without FRM\&FFM & 42.14       & 59.00  & 56.54~\obf{-9.76}  \\

\bottomrule
\end{tabular}
}
}
\end{table}

%% file: Tables/compare_attn.tex
\begin{table}
\centering
\caption{Comparison of convolution-, pooling- and self-attention blocks in the RGB- and accompanying branch, respectively.}
\label{tab:compare_attn}
\resizebox{\columnwidth}{!}{
\renewcommand{\arraystretch}{1.}
\setlength{\tabcolsep}{1mm}{
\begin{tabular}{ll|cc|c} 
\toprule
\textbf{RGB} &\textbf{Accompanying} & \multirow{2}{*}{\textbf{\#Params(M)}} & \multirow{2}{*}{\textbf{GFLOPs}} & \multirow{2}{*}{\textbf{mIoU(\%)}} \\
\textbf{Branch} &\textbf{Branch} & & &\\\midrule\hline
\multirow{7}{*}{MHSA~\cite{xie2021segformer}+}&\cellcolor{gray!10}MHSA~\cite{xie2021segformer}   & 66.87       & 68.39  & 62.92  \\
&\cellcolor{gray!10}ConvNeXt~\cite{liu2022convnext} & 56.42      & 59.85 & 63.73 \\
&\cellcolor{gray!10}FAN~\cite{zhou2022fan} & 68.10 & 69.49 & 63.73\\
&\cellcolor{gray!10}PoolFormer~\cite{yu2022metaformer} & \textbf{56.22} & \textbf{59.52} & 63.83 \\
&\cellcolor{gray!10}$g^n$Conv~\cite{rao2022hornet}      & 62.04       & 64.83  & 64.06  \\
&\cellcolor{gray!10}MSCA~\cite{guo2022segnext} & 61.95       & 68.42  & 64.71  \\
&\cellcolor{gray!10}P2T~\cite{wu2022p2t} & 63.01 & 71.13 & 65.13\\
&\cellcolor{gray!10}PPX (ours) & 58.73 & 65.42 & \textbf{66.30} \\\hdashline
\cellcolor{gray!10}PPX &\multirow{3}{*}{+PPX (ours)} & \textbf{50.88} & 62.95 & 62.21\\
\cellcolor{gray!10}MSCA~\cite{guo2022segnext}&& 62.42 & \textbf{61.10} & 62.88 \\
\cellcolor{gray!10}MHSA~\cite{xie2021segformer}& & 58.73 & 65.42 & \textbf{66.30} \\
\bottomrule
\end{tabular}
}
}
\end{table}

%% file: Tex_content/conclusion.tex
In this work, we tackle arbitrary-modal semantic segmentation.
We put forward the \textsc{DeLiVER} multimodal dataset with four modalities and partial sensor failures under various weather conditions.
We propose the \emph{Hub2Fuse} paradigm with asymmetric branches and design a universal model \emph{CMNeXt} for arbitrary-modal fusion with Self-Query Hub (SQ-Hub) to dynamically select complementary representations and Parallel Pooling Mixer (PPX) to efficiently and flexibly harvest discriminative cross-modal features.
Our CMNeXt sets the new state of the art on six datasets, which can scale from $1$ to $81$ modalities.

\noindent\textbf{Limitations.}
Our asymmetric architecture leverages the assumption that the RGB representation is essential for semantic segmentation, which is partially due to the fact that most pretrained weights are learned on RGB image datasets. Thus, multi-modal pretraining could be beneficial to further improve the flexibility in arbitrary-modal segmentation.
Besides, while the \textsc{DeLiVER} dataset provides multi-view data and instance labels, only the front-view and semantics are exploited in this work.
Aside from these, the fusion of 3D representations of LiDAR and Event data could be addressed in our future work based on the \textsc{DeLiVER} dataset.

%% file: Tex_content/appendix.tex
\clearpage
\appendix
\section{\textsc{DeLiVER} Dataset}

\subsection{Detailed settings in data collection}
\noindent\textbf{Depth2Frames.} The depth camera straightforwardly outputs a grayscale depth map (\ie $0$--$255$ scales), which will cause discontinuity and quantization errors in distance measurements. Therefore, we convert the original depth image to the depth frame using a logarithmic scale, leading to milimetric granularity and better precision at close ranges.

\noindent\textbf{Event2Frames.} The positive- and negative event threshold of the event camera are both set to $0.3$. We record raw event point cloud between two adjacent frames and convert the last occurring event among all pixels into an event frame, where blue indicates positive and red indicates negative.

\noindent\textbf{LiDAR2Frames.} We transform the LiDAR point cloud to the image coordinate system, so as to obtain an image-like representation of LiDAR data. The Field-of-View (FoV) of the front camera is $91^\circ$ and the image resolution is $H{\times}W{=}1042{\times}1042$. The origin is  $(u_0, v_0){=}(H/2, W/2)$. The focal length $(f_x, f_y)$ is calculated as:
\begin{align}
    f_x{=}H/(2{\times}tan(FoV{\times}\pi/360)),\\
    f_y{=}W/(2{\times}tan(FoV{\times}\pi/360)).
\end{align}

To project 3D points to 2D image coordinate, we have:
\begin{align}
\begin{bmatrix}
u \\
v \\
1 \\
\end{bmatrix} = 
\begin{bmatrix}
f_x & 0 & u_0 \\
0 & f_y & v_0 \\
0 & 0 & 1 \\
\end{bmatrix} 
\begin{bmatrix}
\boldsymbol{R} & \boldsymbol{t} \\
\boldsymbol{0}^T_{3{\times}1} & 1 \\
\end{bmatrix}
\begin{bmatrix}
X \\
Y \\
Z \\
1 \\
\end{bmatrix},
\end{align}
where $(X,Y,Z)$ is the LiDAR point, $(u, v)$ is the 2D image pixel, and the rotation ($\boldsymbol{R}$) and the translation ($\boldsymbol{t}$) matrices are set as the unit matrix in the CARLA simulator~\cite{dosovitskiy2017carla}.

\begin{figure*}
    \centering
    \includegraphics[width=1\textwidth]{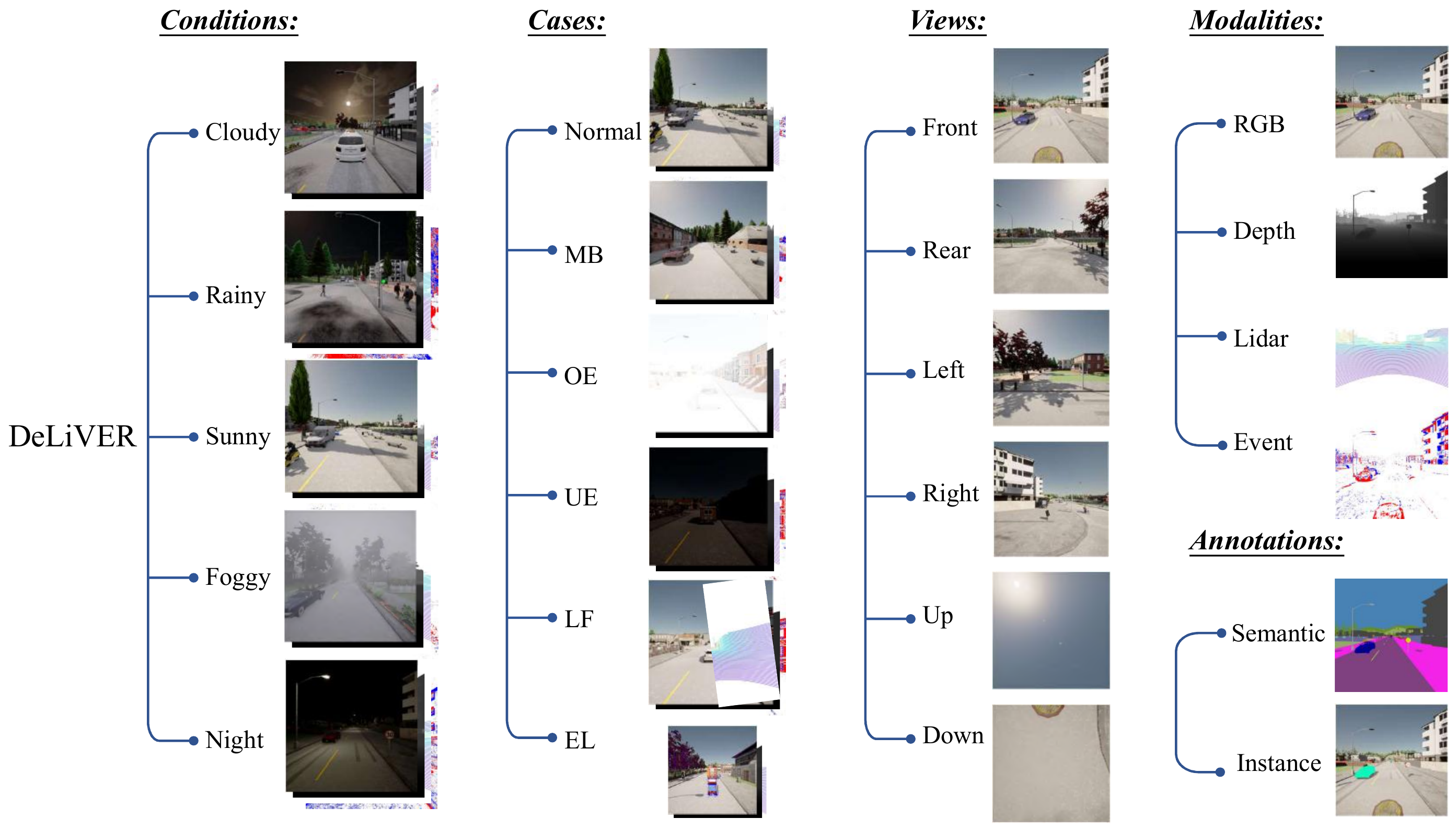}
    \caption{Data structure of the \textsc{DeLiVER} dataset. The columns from left to right are respective conditions, cases, multiple views, modalities and annotations. \textbf{MB}: Motion Blur; \textbf{OE}: Over-Exposure; \textbf{UE}: Under-Exposure; \textbf{LJ}: LiDAR-Jitter; and \textbf{EL}: Event Low-resolution.}
    \label{fig:deliver_struc}
\end{figure*}
\subsection{Dataset structure}
\textsc{DeLiVER} contains Depth, LiDAR, Event, and RGB modalities.
As shown in Fig.~\ref{fig:deliver_struc}, four adverse road scene conditions of \emph{rainy}, \emph{sunny}, \emph{foggy}, and \emph{night} are included in our dataset.
There are five sensor failure cases including Motion Blur (\textbf{MB}), Over-Exposure (\textbf{OE}), Under-Exposure (\textbf{UE}), LiDAR-Jitter (\textbf{LJ}), and Event Low-resolution (\textbf{EL}) to verify that the performance of model is robust and stable in the presence of sensor failures. The sensors are mounted at different locations on the ego car to provide multiple views including \emph{front}, \emph{rear}, \emph{left}, \emph{right}, \emph{up}, and \emph{down}. Each sample is annotated with semantic and instance labels. In this work, we focus on the front-view semantic segmentation.

The $25$ semantic classes in \textsc{DeLiVER} dataset are: \emph{Building, Fence, Other, Pedestrian, Pole, RoadLine, Road, SideWalk, Vegetation, Cars, Wall, TrafficSign, Sky, Ground, Bridge, RailTrack, GroundRail, TrafficLight, Static, Dynamic, Water, Terrain, TwoWheeler, Bus, Truck}.

\subsection{Dataset statistics}
\input{Tables/sup_deliver_stat}
We present statistics of the \textsc{DeLiVER} dataset in Table~\ref{tab:sup_deliver_stat}. We discuss data partitioning in two groups, one according to the conditions and the other according to the sensor failures. Note that, the two groups are mutually inclusive. The five cases from the second group are included in each of five conditions from the first group. For example, cases of \textbf{MB}, \textbf{OE}, \textbf{UE}, \textbf{LJ}, and \textbf{EL} are included in \textit{cloudy}, \textit{foggy}, \textit{night}, \textit{rainy}, and \textit{sunny} conditions, but with different samples. To investigate the robustness under sensor failures, we collect $1199$, $400$, $398$, $398$, and $409$ frames on respective cases.

\subsection{Dataset comparison}
\input{Tables/sup_dataset_comp}
As shown in Table~\ref{tab:sup_dataset_comp}, we compare several datasets with adverse conditions and cases.
All the datasets cover the whole daytime.
The real-scene datasets, \eg, WildDash~\cite{Zendel2018WildDashC} and Waymo\cite{Sun2020waymo}, capture data by using only one or a few sensors, which results a lack of data diversity. 
In contrast, our \textsc{DeLiVER} dataset has four different modalities, including \textit{RGB}, \textit{Depth}, \textit{Event} and \textit{LiDAR}, which enables the multimodal semantic segmentation task to involve up to $4$ modalities. 
Compared to previous synthetic datasets, \eg, SELMA~\cite{testolina2022selma}, SynWoodScape~\cite{sekkat2022synwoodscape}, SynPASS~\cite{zhang2022trans4pass}, our \textsc{DeLiVER} additionally includes $5$ types of sensor failure.
Each sample has semantic and instance annotations, so semantic, instance and panoptic segmentation tasks can be conducted on our \textsc{DeLiVER} dataset. 

\section{Implementation Details}
We conduct our experiments with PyTorch $1.9.0$. All models are trained on a node with 4 A100 GPUs. Below we describe the specific implementation details for six datasets. 

\noindent\textbf{Data representation.}
For depth images, we follow SA-Gate~\cite{chen2020sagate} and CMX~\cite{liu2022cmx} to preprocess the one-channel depth images to HHA-encoded representations~\cite{gupta2014learning}, where HHA includes horizontal disparity, height above ground, and norm angle. The 3D LiDAR and Event data of \textsc{DeLiVER} dataset are transformed to the aforementioned frame format. Then, both LiDAR- and Event-based data are preprocessed as 2D range views~\cite{zhuang2021pmf} and 3-channel representations~\cite{zhang2021issafe}, respectively. 

\noindent\textbf{\textsc{DeLiVER} dataset.} 
We train our models for $200$ epochs on the \textsc{DeLiVER} dataset. The batch size is $2$ on each of four GPUs. The resolution of all modalities is set as $1024{\times}1024$ for training and inference. In the Event Low-resolution cases, the Event-based images with the original size of $260{\times}260$ are upsampled to $1024{\times}1024$. During evaluation, we only apply the single-scale test strategy. The backbone of CMNeXt is based on MiT-B2~\cite{xie2021segformer}. To verify the effectiveness of our method under convolutional networks, the CNN-based SegNeXt-Base~\cite{guo2022segnext} is selected as the backbone, when compared to the MiT-B2 one.

\noindent\textbf{KITTI-360 dataset.} As there are more than $49K$ training data on KITTI-360 dataset, the models are trained for $40$ epochs. The image resolution is set as $1408{\times}376$ and the batch size is $4$ on each of four GPUs. The backbone of CMNeXt is based on MiT-B2~\cite{xie2021segformer}. 

\noindent\textbf{NYU Depth V2 dataset.} Following CMX~\cite{liu2022cmx}, the number of training epochs is set as $500$ for a fair comparison. The resolution of RGB and Depth images is set as $640{\times}480$. The training batch size is $4$ on each of four GPUs. The backbone of CMNeXt is based on MiT-B4~\cite{xie2021segformer}. We apply the multi-scale flip test strategy for a fair comparison.

\noindent\textbf{MFNet dataset.} We train our CMNeXt models with the MiT-B4 backbone for $500$ epochs on the MFNet dataset. The resolution of RGB and Thermal images is set as $640{\times}480$ and the batch size is $4$ on each of four GPUs. We apply the multi-scale flip test strategy for a fair comparison.

\noindent\textbf{MCubeS dataset.} To compare with MCubeSNet~\cite{liang2022mcubesnet}, we build CMNeXt with MiT-B2 and train the model for $500$ epochs. Following MCubeSNet~\cite{liang2022mcubesnet}, the image size is set as $512{\times}512$ during training and $1024{\times}1024$ during evaluation. The batch size is set as $4$ on each of four GPUs. 

\noindent\textbf{UrbanLF dataset.} To perform comparison with the OCR-LF model~\cite{sheng2022urbanlf}, we build CMNeXt with MiT-B4. The image size on the real and synthetic sets is $640{\times}480$. The angular resolution of $81$ sub-aperture images of the UrbanLF dataset is $9{\times}9$. To conduct arbitrary-modal segmentation, the center-aperture image is selected as the primary modality, while the other apertures are as additional modalities. We sample respective $8$, $33$, and $80$ light field images as the supplementary modalities, \ie, LF$8$, LF$33$, and LF$80$ for short. The $8$ images are from the center horizontal direction, while the $33$ images are from the four directions of horizontal, vertical, $\frac{1}{4}\pi$, and $\frac{3}{4}\pi$, following UrbanLF~\cite{sheng2022urbanlf}. 

\begin{figure}[t]
    \centering
    \includegraphics[width=1\columnwidth]{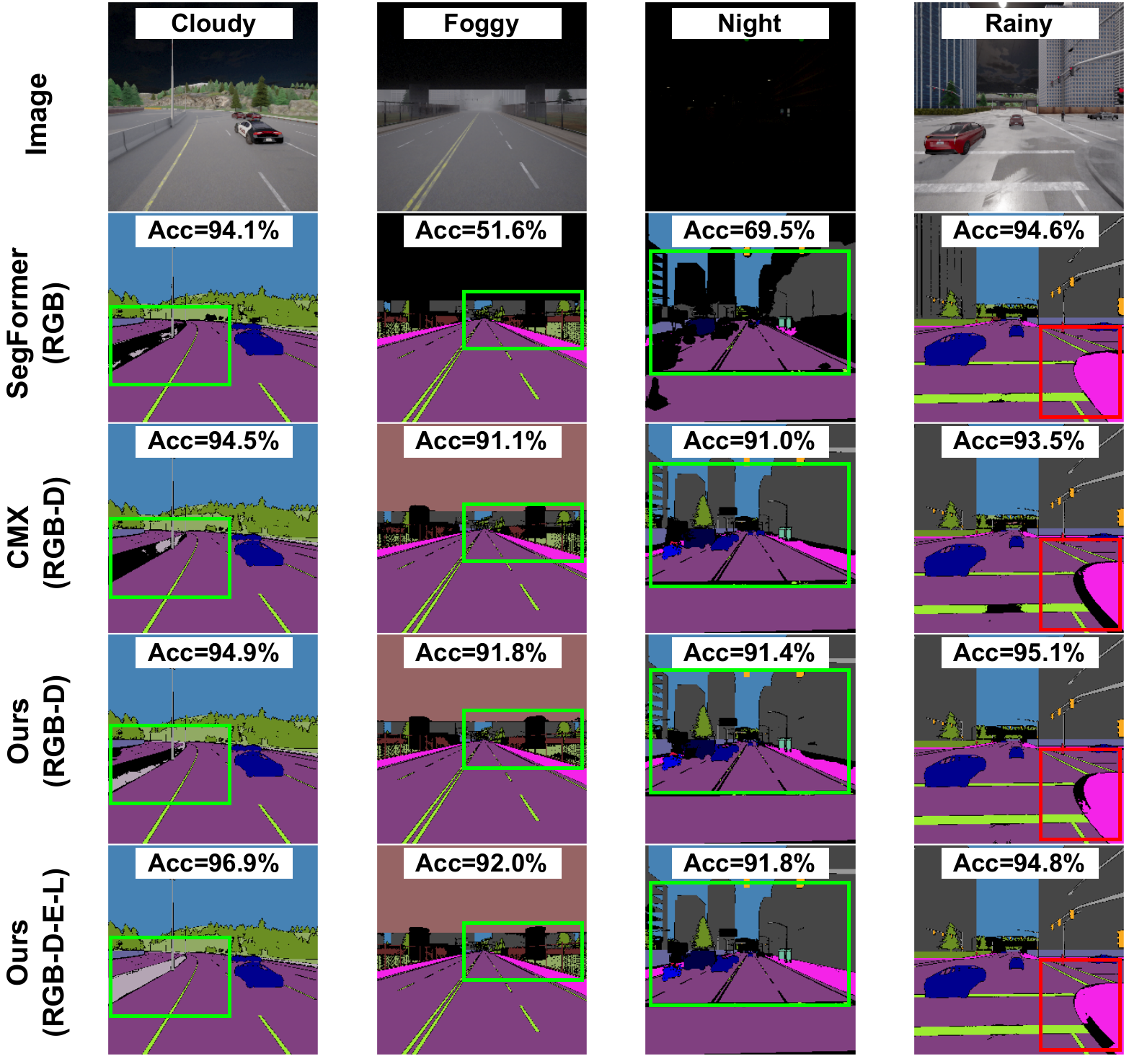}
    \vskip -1ex
    \caption{More visualization results on \textsc{DeLiVER} dataset. From left to right are the respective \textit{cloudy}, \textit{foggy}, \textit{night} and \textit{rainy} scene.}
    \label{fig:sup_qualitative_vis}
    \vskip -2ex
\end{figure}
\section{More visualizations on \textsc{DeLiVER}}
As shown in Fig.~\ref{fig:sup_qualitative_vis}, in the four adverse weather conditions, RGB-D fusion-based methods greatly improve the performance, particularly for distant elements in \textit{foggy} and \textit{nighttime} scenes.
Our RGB-D solution is more accurate than CMX (RGB-D),
and the full quad-modal RGB-D-E-L CMNeXt model further enhances the segmentation. A failure case is shown on the right column (\ie, the \textit{rainy} scene) of Fig.~\ref{fig:sup_qualitative_vis}, in which the RGB-only model has a better segmentation on the \textit{sidewalk} class. However, our quad-modal CMNeXt has a higher accuracy score with $94.8\%$.

\section{Acknowledgments}
This work was supported in part by Helmholtz Association of German Research Centers, in part by the Federal Ministry of Labor and Social Affairs (BMAS) through the AccessibleMaps project under Grant 01KM151112, in part by the University of Excellence through the ``KIT Future Fields'' project, and in part by Hangzhou SurImage Technology Company Ltd. This work was partially performed on the HoreKa supercomputer funded by the Ministry of Science, Research and the Arts Baden-Württemberg and by the Federal Ministry of Education and Research.

%% file: Tables/sup_deliver_stat.tex
\begin{table*}
\centering
\caption{Data statistic of DeLiVER dataset. It includes four adverse conditions (\emph{cloudy}, \emph{foggy}, \emph{rainy}, and \emph{night}), and each condition has five failure cases (\textbf{MB}: Motion Blur; \textbf{OE}: Over-Exposure; \textbf{UE}: Under-Exposure; \textbf{LJ}: LiDAR-Jitter; and \textbf{EL}: Event Low-resolution).}
\label{tab:sup_deliver_stat}
\resizebox{\textwidth}{!}{
\setlength{\tabcolsep}{3mm}{
\begin{tabular}{l|rrrrr:r||rrrrrr:r} 
\toprule
\textbf{Split} & \textbf{Cloudy} & \textbf{Foggy} & \textbf{Night} & \textbf{Rainny} & \textbf{Sunny} & \textbf{Total} & \textbf{Normal} & \textbf{MB} & \textbf{OE} & \textbf{UE} & \textbf{LJ} & \textbf{EL} & \textbf{Total}  \\\midrule\midrule
Train & 794 & 795 & 797 & 799 & 798 & 3983  & 2585 & 600 & 200 & 199 & 199 & 200 & 3983 \\
Val & 398 & 400 & 410 & 398 & 399 & 2005 & 1298 & 299 & 100 & 99 & 100 & 109 & 2005 \\
Test & 379 & 379 & 379 & 380 & 380 & 1897 & 1198 & 300 & 100 & 100 & 99 & 100 & 1897 \\\hline
Front-view & 1571 & 1574 & 1586 & 1577 & 1577 & 7885 & 5081 & 1199 & 400 & 398 & 398 & 409 & 7885 \\\hline
All six views & 9426 & 9444 & 9516 & 9462 & 9462 & 47310 & 30486 & 7194 & 2400 & 2388 & 2388 & 2454 & 47310 \\
\bottomrule
\end{tabular}
}
}
\end{table*}

%% file: Tables/sup_dataset_comp.tex
\begin{table*}[!t]
\centering
\caption{
Comparison between multimodal datasets. D:Day; S:Sunset; N:Night; *:random; Sem.:Semantic; Ins.:Instance.
}
\label{tab:sup_dataset_comp}
\resizebox{\textwidth}{!}{
\setlength{\tabcolsep}{3mm}{
\begin{tabular}{c|c:cccc:c:c:ccc:c:cc}
\toprule
    \multicolumn{1}{c|}{\multirow{2}{*}{Dataset}}&\multirow{2}{*}{Type}&\multicolumn{4}{c:}{Sensors}&\multicolumn{1}{c:}{Sensor}&\multicolumn{1}{c:}{RGB}&\multicolumn{3}{c:}{Diversity} &\multirow{2}{*}{Classes}&\multicolumn{2}{c}{Labels}\\
    &&Camera&Depth&Event&LiDAR&Failures&Failures&Weathers&Daytime&Views&&Sem.&Ins.\\
\midrule
\midrule
WildDash~\cite{Zendel2018WildDashC}&Real&1&0&0&0&0&15&*&*&*&19&$\checkmark$&$\checkmark$\\
Waymo~\cite{Sun2020waymo}&Real&5&0&0&5&0&0&2&DN&5&28&$\checkmark$&$\checkmark$\\
SELMA~\cite{testolina2022selma}&Synthetic&7&7&0&3&0&6&9&DSN&7&19&$\checkmark$&$\times$\\
SynWoodScape~\cite{sekkat2022synwoodscape}&Synthetic&5&5&5&1&0&0&4&DS&5&25&$\checkmark$&$\checkmark$\\
SynPASS~\cite{zhang2022trans4pass}&Synthetic&6&0&0&0&0&0&4&DN&1&22&$\checkmark$&$\times$\\
\rowcolor{gray!15}DeLiVER (ours)&Synthetic&6&6&6&1&5&3&4&DN&6&25&$\checkmark$&$\checkmark$\\
\bottomrule

\end{tabular}
}
}
\end{table*}